\pdfoutput=1

\documentclass[11pt]{article}

\usepackage{ACL2023}

\usepackage{times}
\usepackage{latexsym}

\usepackage[T1]{fontenc}

\usepackage[utf8]{inputenc}

\usepackage{microtype}

\usepackage{inconsolata}

\usepackage{booktabs}
\usepackage{graphicx}
\usepackage{xspace}
\usepackage{multirow}
\usepackage{colortbl}
\usepackage{mdframed}
\usepackage{tcolorbox}
\usepackage{longtable}
\usepackage{amssymb}
\usepackage{amsbsy}
\usepackage{bbding}

\newcommand{\rw}[1]{\textcolor[rgb]{0.00,0.00,0.00}{#1}} %

\newcommand{\OurDataset}{\textsc{Fallacies}\xspace}

\title{A Closer Look at the Self-Verification Abilities of \\Large Language Models in Logical Reasoning}

\author{Ruixin Hong\textsuperscript{1,2}\thanks{~~~Work done during the internship at Tencent AI Lab.}, 
Hongming Zhang\textsuperscript{2}, 
Xinyu Pang\textsuperscript{1}, 
Dong Yu\textsuperscript{2}, 
Changshui Zhang\textsuperscript{1} \\
\textsuperscript{1}Institute for Artificial Intelligence, Tsinghua University (THUAI); \\
\textsuperscript{1}Beijing National Research Center for Information Science and Technology (BNRist); \\
\textsuperscript{1}Department of Automation, Tsinghua University, Beijing, P.R.China \\
\textsuperscript{2}Tencent AI Lab, Seattle \\
\texttt{hrx20@mails.tsinghua.edu.cn,} 
\texttt{hongmzhang@tencent.com,} \\
}

\begin{document}
\maketitle
\begin{abstract}
Logical reasoning has been an ongoing pursuit in the field of AI. 
Despite significant advancements made by large language models (LLMs), they still struggle with complex logical reasoning problems.
To enhance reasoning performance, one promising direction is scalable oversight, which requires LLMs to identify their own errors and then improve by themselves.
Various self-verification methods have been proposed in pursuit of this goal. 
Nevertheless, whether existing models understand their own errors well is still under investigation.
In this paper, we take a closer look at the self-verification abilities of LLMs in the context of logical reasoning, focusing on their ability to identify logical fallacies accurately. 
We introduce a dataset, \OurDataset, containing 232 types of reasoning fallacies categorized in a hierarchical taxonomy.
By conducting exhaustive experiments on \OurDataset, we obtain comprehensive and detailed analyses of a series of models on their verification abilities.
Our main findings suggest that existing LLMs could struggle to identify fallacious reasoning steps accurately and may fall short of guaranteeing the validity of self-verification methods.
Drawing from these observations, we offer suggestions for future research and practical applications of self-verification methods.\footnote{Data is available at \url{https://github.com/Raising-hrx/FALLACIES}.
}
\end{abstract}

\begin{figure}[t!]
  \centering
  \includegraphics[width=\linewidth]{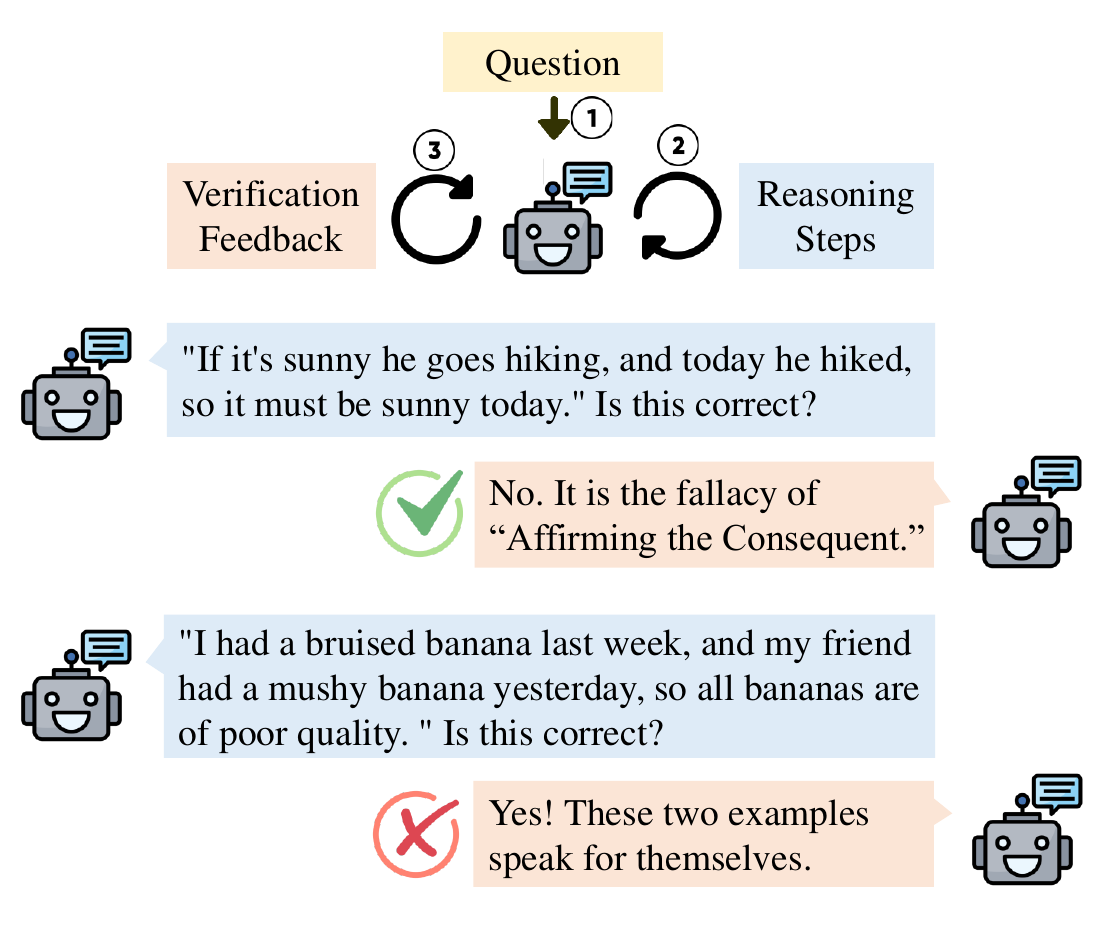}
  \caption{The self-verification approach requires LLMs to identify the fallacious steps in their own reasoning process. However, LLMs might be susceptible to certain types of fallacies and fail to identify them, leading to the potential failure of self-verification.}
  \vspace{-5mm}
  \label{fig:intro}
\end{figure}

\section{Introduction}

Logical reasoning is not only a crucial aspect of human intelligence but also one of the long-term pursuits of artificial intelligence~\cite{mccarthy1989artificial}. 
It is indispensable in intelligent systems, enabling problem-solving, decision-making, and critical thinking.
Large language models (LLMs) have recently achieved remarkable advancements in a wide range of tasks~\cite{DBLP:journals/corr/abs-2303-08774}.
Being prompted appropriately, LLMs exhibit the emergent ability to ``reason'' step by step like humans~\cite{DBLP:journals/tmlr/WeiTBRZBYBZMCHVLDF22,DBLP:conf/nips/Wei0SBIXCLZ22}.
Nonetheless, increasing research suggests that LLMs struggle with intricate logical reasoning problems, occasionally producing unfaithful reasoning steps fraught with logical fallacies~\cite{DBLP:journals/corr/abs-2308-03762,DBLP:journals/corr/abs-2304-03439,DBLP:journals/corr/abs-2305-12295,DBLP:conf/acl/HongZZ0Z23}.

To tackle this issue, a prevalent and promising approach is scalable oversight~\cite{DBLP:journals/corr/abs-2211-03540}, where the LLMs could be boosted based on their own evaluation signals~\cite{superalignment}.
In this regard, various strategies of self-verification using LLMs are proposed to enhance reasoning performance~\cite{DBLP:journals/corr/abs-2306-03872,DBLP:journals/corr/abs-2212-09561,DBLP:journals/corr/abs-2305-00633,DBLP:journals/corr/abs-2303-17651,DBLP:journals/corr/abs-2308-00436}.
As shown in Figure~\ref{fig:intro}, the LLMs first generate the reasoning process and then self-verify their own output. 
The verification results are then used to refine the output or further improve the models. 
Their \textit{assumption} is that LLMs can reliably identify fallacious reasoning steps.

Though empirical evidence demonstrates their preliminary success, a thorough and comprehensive evaluation of the underlying assumption remains unexplored.
First, these efforts typically use the performance of the final task (e.g., answer accuracy) to illustrate their effectiveness. 
However, this is a proxy metric that does not directly reflect the ability of LLMs to identify logical fallacies.
LLMs might possibly arrive at the correct answer despite the existence of fallacious intermediate steps~\cite{DBLP:conf/nips/Wei0SBIXCLZ22,DBLP:conf/iclr/CreswellSH23,DBLP:journals/corr/abs-2307-13702}. 
Second, they are usually only concerned with \textit{whether or not} the reasoning step is fallacious, rather than \textit{what type of} fallacy exists in the step, which could be informative for the self-improvement of LLMs.
Such an oversimplification precludes a definitive analysis of the LLMs' ability to identify different types of fallacies.

In this paper, we provide a comprehensive evaluation of the verification abilities of LLMs in logical reasoning.
Specifically, we collect a dataset, \OurDataset, containing 4,640 reasoning steps for 232 types of fallacies.
We evaluate whether LLMs can distinguish between correct and fallacious reasoning steps.
Such directed evaluation can provide a more accurate reflection of the verification abilities of LLMs, as it steers away from proxy metrics and delves straight into the actual performance in identifying fallacious steps. 
Compared to previous datasets, our dataset features more types of fallacies, a larger scale, more fine-grained reasoning, and explicit premises and conclusions.
Furthermore, we adopt a hierarchical taxonomy of fallacies, which divides fallacies into two main categories and nine subcategories, allowing for a more systematic approach toward analyzing the verification abilities of LLMs across varying aspects.

We conduct exhaustive experiments on a range of LLMs.
First, experimental results show that most LLMs struggle with accurately identifying the fallacious steps.
Most LLMs only achieved an overall accuracy rate of less than 80\%, suggesting that LLMs could lack sufficient logical verification abilities.
Thus, we should be more cautious about the self-verification methods of LLMs.
Second, the performance of LLMs can be remarkably imbalanced in different types of fallacies.
Most LLMs perform much worse at identifying fallacies related to logical structure than those related to content, 
pointing toward key directions for improving LLMs' verification and reasoning abilities.
Third, we also find that LLMs encounter challenges when it comes to classifying different types of fallacies. 
\rw{Presenting LLMs with the definitions of fallacies does not appear to improve their ability to recognize fallacies.}
This raises a call for further research to delve into the underlying mechanisms through which LLMs understand reasoning and fallacies.
In summary, we present a comprehensive evaluation of the verification abilities of LLMs and highlight their limitations in identifying fallacies, 
urging the community to apply the self-verification methods with caution.

\begin{figure*}[t!]
  \centering
  \includegraphics[width=\textwidth]{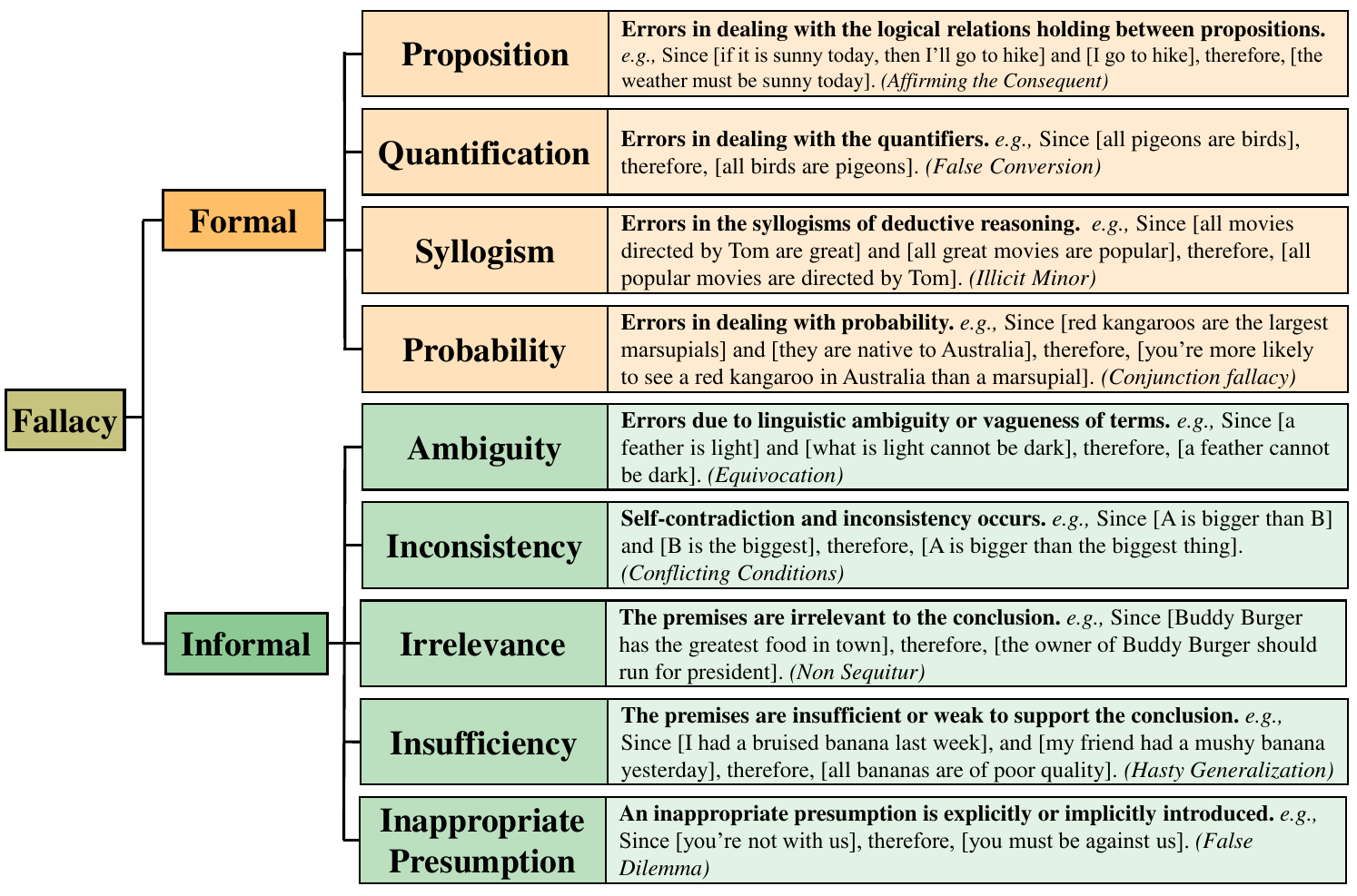}
  \caption{The hierarchical taxonomy of fallacies. For each sub-category, we present its definition and an example of a fallacy within the sub-category. We use square brackets to indicate the premises and conclusions.}
  \label{fig:taxonomy}
  \vspace{2mm}
\end{figure*}

\section{Related Work}
\subsection{Language Models for Logical Reasoning}
Compared with traditional symbol-based logic reasoning systems, using language models to directly reason over natural language is a more flexible and popular approach~\cite{DBLP:journals/corr/abs-2303-14725,DBLP:journals/corr/abs-2303-12023}.
Many efforts have been devoted to improving the logical reasoning abilities of language models from different perspectives, including fine-tuning methods~\cite{DBLP:conf/ijcai/ClarkTR20,DBLP:conf/emnlp/DalviJTXSPC21}, pre-training methods~\cite{DBLP:conf/nips/PiZ0DL22,DBLP:conf/acl/JiaoGSN22}, and modular methods~\cite{DBLP:conf/naacl/HongZYZ22,DBLP:conf/emnlp/Yang0C22}.
As the model scale increases, specially designed prompts (e.g., Chain-of-Thought prompts~\cite{DBLP:conf/nips/Wei0SBIXCLZ22}) can elicit the step-by-step reasoning abilities of LLMs, achieving remarkable improvement in multiple reasoning tasks~\cite{chu2023survey}.
However, some studies find that LLMs still struggle with complex logical reasoning problems~\cite{DBLP:journals/corr/abs-2308-03762} and could be susceptible to logical fallacies~\cite{DBLP:journals/corr/abs-2308-09853}.
There is still a lack of comprehensive research to investigate the understanding of LLMs on different logical fallacies.

\subsection{Self-Verification with Large Language Models}
\rw{A prevalent method to enhance the capacity of LLMs is through learning and correction via high-quality verification feedback. 
This feedback proves instrumental in various aspects of model optimization. 
For instance, it can be used to fine-tune the models~\cite{DBLP:conf/nips/Ouyang0JAWMZASR22,DBLP:journals/corr/abs-2303-16755,DBLP:journals/corr/abs-2303-16749,DBLP:conf/nips/ZelikmanWMG22,DBLP:conf/acl/WangKMLSKH23,DBLP:journals/corr/abs-2212-08073,DBLP:journals/corr/abs-2210-11610}. In addition, verification feedback can be utilized to re-rank the model outputs~\cite{DBLP:journals/corr/abs-2305-20050,DBLP:journals/corr/abs-2212-09561,DBLP:journals/corr/abs-2301-00303,DBLP:conf/icml/Ni0RSYWL23,DBLP:journals/corr/abs-2306-03872}. 
It can also be employed to refine the outputs~\cite{DBLP:journals/corr/abs-2303-17651,Shinn2023ReflexionLA,DBLP:journals/corr/abs-2306-03856} and guide the generation process of the models~\cite{DBLP:journals/corr/abs-2305-10601,DBLP:journals/corr/abs-2305-00633}.}
A straightforward approach to collecting feedback is to collect directly from humans~\cite{DBLP:conf/nips/Ouyang0JAWMZASR22,DBLP:journals/corr/abs-2305-00955}, but this could prove to be costly and unable to provide instant feedback.
Alternatively, collecting feedback from external tools~\cite{DBLP:journals/corr/abs-2305-11738,DBLP:journals/corr/abs-2307-13528} or metrics~\cite{DBLP:conf/emnlp/JungQWBB0C22} could be more feasible but limited to specific tasks.
Thus, some researchers turn to use large language models themselves to provide verification feedback~\cite{DBLP:journals/corr/abs-2306-03872,DBLP:journals/corr/abs-2212-09561,DBLP:journals/corr/abs-2305-00633,DBLP:journals/corr/abs-2303-17651,DBLP:journals/corr/abs-2308-00436}, which is more scalable.
Nonetheless, recent papers~\cite{huang2023large,DBLP:journals/corr/abs-2310-08118} have raised doubts about the self-verification abilities of the LLMs.
\rw{For instance, \citet{huang2023large} find that LLMs struggle to self-correct their responses without external feedback.}
However, they still leave some open questions for subsequent research, such as what exactly the performance of LLMs is to verify a single reasoning step.
In this paper, we delve deeply into this subject, critically examining the verification abilities of LLMs from the perspective of logical reasoning.

\begin{table}[t]
\small
\centering
\begin{tabular}{@{}lc@{}}
\toprule
\textbf{Category} & \textbf{\# Fallacy} \\ \midrule
\textbf{Formal} & \textbf{24} \\
\quad proposition (prop.) & 6 \\
\quad quantification (quant.)& 6 \\ 
\quad syllogism (syl.) & 8 \\
\quad probability (prob.) & 4 \\
\midrule
\textbf{Informal} & \textbf{208} \\
\quad ambiguity (amb.) & 15 \\
\quad inconsistency (incon.) & 3 \\
\quad irrelevance (irrel.) & 78 \\
\quad insufficiency (insuf.) & 58 \\
\quad inappropriate presumption (inappr.) & 54 \\ \bottomrule
\end{tabular}%
\caption{Distribution of 232 fallacies in \OurDataset. }
\label{tab:fallacy_distribution}
\end{table}

\section{\OurDataset}
This section outlines the design principles and process of constructing our dataset \OurDataset.

\subsection{Design Principles}
\noindent \textbf{Covering more comprehensive error types:}
We propose to evaluate the verification abilities of LLMs on a wider range of types of errors.
Relying solely on a single existing dataset of logical reasoning might pose a challenge in drawing comprehensive conclusions, as it may struggle to cover the types of errors that can occur in realistic scenarios.
For example, synthetic datasets (e.g., ProofWriter~\cite{DBLP:conf/acl/TafjordDC21}) are generated from fixed logic templates and vocabularies, which could leave out the reasoning errors caused by verbal ambiguity.

\noindent \textbf{Hierarchical fallacy taxonomy:} 
We propose to categorize reasoning errors at a fine-grained, hierarchical level. 
This approach enables a more thorough assessment of the performance of LLMs across various types of fallacies and offers a more comprehensive perspective on their effectiveness.
Meanwhile, we ensure that each reasoning step exclusively pertains to a single type of fallacy to prevent different types of fallacies from potentially confounding one another.

\noindent \textbf{Clarifying the premises and conclusions:}
The third principle is to indicate the premises and conclusions of the reasoning step explicitly. Reasoning is the process of concluding based on known information, and the validity of reasoning cannot be properly judged without providing premise information. 
Factors like linguistic omissions and implied background knowledge can complicate reasoning in natural language. 
Consequently, in our dataset, we strive to ensure that each step is a complete unit of reasoning that contains enough information to verify its correctness.

\subsection{Taxonomy of Fallacy}
\label{sec:taxonomy_of_fallacy}
A fallacy is an error in reasoning~\cite{bennett2012logically,jevons1872elementary}.
It focuses more on whether the conclusion of an argument can be logically deduced from the premises rather than on factual errors or other aspects~\cite{lau2011introduction}.
Identifying logical fallacies is essential for a judicious reasoning system. 
Without the ability to identify logical fallacies, a reasoning system may lack essential critical thinking skills, leaving it susceptible to illogical arguments and deliberate manipulation.

Fallacies can be classified into two primary categories: \textbf{Formal Fallacies}, errors due to invalid logical structures or inference patterns, and \textbf{Informal Fallacies}, errors due to the content of premises and conclusions.
Based on specific causes of error, we divide each main category into several subcategories.
Each subcategory contains several finest-grained fallacies, divided based on more detailed causes of errors.
Figure~\ref{fig:taxonomy} demonstrates our hierarchical taxonomy of fallacies.
We derive this taxonomy by integrating multiple scholarly resources~\cite{fieser2011internet, magnus2005forall, britannica-fallacy}.

\subsection{Data Collection}
\label{sec:data_collection}

We first collect 232 types of fallacies from classic academic books~\cite{bennett2012logically,fieser2011internet}. 
\rw{The authors of these books carefully collect these fallacies from many available academic resources (including peer-reviewed journals, encyclopedias, and books), covering a substantial portion of common errors in logical reasoning.}
We collect the definition of each fallacy, i.e., an article containing the description and examples of that fallacy.
These definitions are used in academic books to help humans understand the meaning of fallacies.
Figure~\ref{fig:fallacy_definition} in the Appendix shows an example of the definition.
We then assign these 232 fallacies to appropriate categories according to the taxonomy in Sec.~\ref{sec:taxonomy_of_fallacy}.
Table~\ref{tab:fallacy_distribution} shows the distribution of fallacies.
Detailed categorizations and descriptions of these fallacies are in Appendix Table~\ref{tab:fallacy_232}.

Then, we collect fallacious and correct reasoning steps.
Considering that directly creating fallacious steps from scratch can be challenging, we adopt a strategy in which powerful LLMs first generate candidates, and then we let human experts revise them.
Specifically, we ask GPT-4~\cite{DBLP:journals/corr/abs-2303-08774} to generate fallacious steps based on the collected definitions of fallacies. 
To generate diverse steps covering a wider range of domains, we explicitly require the model to generate around a theme, a randomly sampled noun from ConceptNet~\cite{DBLP:conf/aaai/SpeerCH17}.
Detailed prompt can be found in Appendix~\ref{sec:appendix_data}.
Subsequently, human experts carefully proofread and refine these candidates to ensure they fall within the corresponding fallacies.
Meanwhile, we require experts to make each step a single inference~\cite{DBLP:series/synthesis/2013Dagan}, rather than complex reasoning that involves multiple intermediate steps.

\begin{table*}[t]
\centering
\resizebox{\textwidth}{!}{%
\begin{tabular}{@{}lcccccc@{}}
\toprule
\multicolumn{1}{@{}l@{}}{\textbf{Dataset}} & \begin{tabular}[c]{@{}c@{}} \textbf{Number} \\ \textbf{of Fallacies}\end{tabular} & \begin{tabular}[c]{@{}c@{}} \textbf{Number} \\ \textbf{of Steps}\end{tabular}  & \begin{tabular}[c]{@{}c@{}} \textbf{Taxonomy} \\ \textbf{of Fallacy}\end{tabular} & \begin{tabular}[c]{@{}c@{}}\textbf{Granularity} \\ \textbf{of Reasoning}\end{tabular} & \begin{tabular}[c]{@{}c@{}}\textbf{Explicit Premises} \\ \textbf{and Conclusions}\end{tabular} & \begin{tabular}[c]{@{}c@{}}\textbf{Identifying Fallacy} \\ \textbf{from Correct Reasoning}\end{tabular} \\ \midrule
\citet{stab-gurevych-2017-recognizing} & 1 & 1,029 & No & Coarse & No & \textbf{Yes} \\
\citet{habernal-etal-2018-name} & 1 & 2,085 & No & Coarse & No & \textbf{Yes} \\
\citet{jin-etal-2022-logical} & 13 & 2,449 & Coarse & Coarse & No & No \\ \midrule
\textbf{\OurDataset (Ours)} & \textbf{232} & \textbf{4,640} & \textbf{Fine \& Hierarchical} & \textbf{Fine} & \textbf{Yes} & \textbf{Yes} \\ \bottomrule
\end{tabular}
}
\caption{Comparison of \OurDataset with existing fallacy-related datasets.}
\label{tab:comparison_dataset}
\end{table*}

We require the experts to fix the fallacious steps to collect the correct contrastive steps.
The experts make as few modifications as possible to turn the fallacious steps into correct ones, which do not contain any reasoning errors.
For instance, the fallacious step of the fallacy of \textit{Affirming the Consequent} in Figure~\ref{fig:taxonomy} can be fixed into \textit{``Since [if it is sunny today, then I’ll go to hike] and [It is sunny today], therefore, [I’ll go to hike today].''}
More contrastive samples can be found in Table~\ref{tab:cases}.

In this way, we obtain ten fallacious and ten correct steps for each of the 232 types of fallacies.
To check data quality, we ask three additional experts to re-annotate 50 randomly sampled steps. 
They annotate each step as a correct step, a fallacious step that belongs to the corresponding fallacy, or a fallacious step that does not belong to the corresponding fallacy.
Their average agreement with the labels achieves 0.856 (Cohen's Kappa), indicating the high quality of our data.

\subsection{Comparison with Existing Dataset}
We compare our dataset and existing fallacy-related datasets in Table~\ref{tab:comparison_dataset}.
\OurDataset holds significant advantages across multiple dimensions.
It encompasses a broader spectrum of fallacy types, presenting a fine-grained and hierarchical taxonomy of fallacies.
Additionally, \OurDataset stands out for its clarity and subtlety, avoiding ambiguous judgments about the correctness of reasoning.
We make explicit the premises and conclusions of each reasoning step and the granularity of reasoning in \OurDataset is finer.
In contrast to the most recent dataset~\cite{jin-etal-2022-logical}, which contained only fallacious reasoning, our dataset includes both correct and fallacious reasoning steps.

\definecolor{mycolor1}{RGB}{247,226,219}
\definecolor{mycolor2}{RGB}{142,141,200}
\definecolor{mycolor3}{RGB}{71,133,90}
\definecolor{mycolor4}{RGB}{255,255,255}

\newcolumntype{X}{>{\columncolor{mycolor1}}c}
\newcolumntype{Y}{>{\columncolor{mycolor2!50}}c}
\newcolumntype{Z}{>{\columncolor{mycolor3!50}}c}
\newcolumntype{K}{>{\columncolor{mycolor4}}c}

\begin{table*}[t!]
\small
\centering
\begin{tabular}{@{}l|XXXXY|XXXXXY|Z}
\toprule
\multirow{2}{*}{\textbf{Model}} & \multicolumn{5}{c|}{\textbf{Formal}} & \multicolumn{6}{c|}{\textbf{Informal}} & \multicolumn{1}{c}{} \\ 
 & \multicolumn{1}{c}{\textbf{prop.}} & \multicolumn{1}{c}{\textbf{quant.}} & \multicolumn{1}{c}{\textbf{syl.}} & \multicolumn{1}{c}{\textbf{prob.}} & \multicolumn{1}{c|}{\textbf{Avg.}} & \multicolumn{1}{c}{\textbf{amb.}} & \multicolumn{1}{c}{\textbf{incon.}} & \multicolumn{1}{c}{\textbf{irrel.}} & \multicolumn{1}{c}{\textbf{insuf.}} & \multicolumn{1}{c}{\textbf{inappr.}} & \multicolumn{1}{c|}{\textbf{Avg.}} & \multicolumn{1}{c}{\multirow{-2}{*}{\textbf{Avg.}}} \\ \midrule
Random &          50.0 &             50.0 &        50.0 &          50.0 &   50.0 &      50.0 &          50.0 &        50.0 &          50.0 &                      50.0 &     50.0 &  50.0 \\
\midrule
Flan-T5-Large &          49.2 &             59.2 &        78.1 &          57.5 &   61.0 &      64.3 &          86.7 &        72.8 &          74.8 &                      69.5 &     73.6 &  67.3 \\
Flan-T5-xl    &          48.3 &             78.3 &        78.8 &          66.3 &   67.9 &      67.7 &          83.3 &        72.8 &          74.5 &                      72.1 &     74.1 &  71.0 \\
Flan-T5-xxl   &          46.7 &             70.8 &        80.6 &          57.5 &   63.9 &      67.0 &          78.3 &        72.6 &          75.7 &                      71.2 &     73.0 &  68.4 \\
Llama2-7B     &          59.2 &             63.3 &        58.8 &          63.7 &   61.3 &      67.7 &          70.0 &        72.7 &          75.3 &                      73.4 &     71.8 &  66.5 \\
Llama2-13B    &          55.8 &             62.5 &        56.2 &          65.0 &   59.9 &      58.3 &          58.3 &        66.9 &          65.1 &                      67.4 &     63.2 &  61.5 \\
Llama2-70B    &          58.3 &             79.2 &        79.4 &          76.2 &   73.3 &      82.0 &          90.0 &        90.3 &          90.4 &                      \textbf{88.2} &     \textbf{88.2} &  80.7 \\
Baichuan2-7B  &          55.8 &             52.5 &        52.5 &          50.0 &   52.7 &      53.3 &          51.7 &        51.7 &          51.9 &                      52.9 &     52.3 &  52.5 \\
Baichuan2-13B &          59.2 &             75.8 &        79.4 &          57.5 &   68.0 &      77.0 &          78.3 &        83.1 &          85.8 &                      80.2 &     80.9 &  74.4 \\
ChatGLM-6B    &          60.0 &             52.5 &        60.6 &          50.0 &   55.8 &      52.3 &          56.7 &        54.7 &          53.0 &                      55.5 &     54.4 &  55.1 \\
ChatGLM2-6B   &          60.8 &             61.7 &        69.4 &          55.0 &   61.7 &      65.3 &          83.3 &        76.7 &          77.9 &                      72.5 &     75.2 &  68.4 \\
InternLM-7B   &          51.7 &             70.0 &        71.9 &          48.8 &   60.6 &      70.7 &          83.3 &        74.9 &          75.9 &                      72.2 &     75.4 &  68.0 \\
InternLM-20B  &          57.5 &             67.5 &        74.4 &          55.0 &   63.6 &      73.0 &          85.0 &        77.0 &          77.8 &                      74.9 &     77.5 &  70.6 \\
Falcon-7B     &          31.7 &             45.8 &        41.9 &          52.5 &   43.0 &      64.7 &          66.7 &        72.9 &          75.3 &                      71.2 &     70.2 &  56.6 \\
WizardLM-13B  &          63.3 &             66.7 &        70.0 &          70.0 &   67.5 &      82.0 &          86.7 &        87.6 &          91.8 &                      84.4 &     86.5 &  77.0 \\
Vicuna-7B     &          65.0 &             75.0 &        73.8 &          60.0 &   68.4 &      77.0 &          83.3 &        84.0 &          85.3 &                      81.7 &     82.2 &  75.3 \\
Vicuna-13B    &          75.0 &             69.2 &        72.5 &          62.5 &   69.8 &      81.0 &          83.3 &        88.9 &          91.5 &                      87.3 &     86.4 &  78.1 \\
Qwen-14B      &          70.0 &             78.3 &        83.1 &          67.5 &   74.7 &      83.0 &          \textbf{91.7} &        88.4 &          91.4 &                      86.6 &     \textbf{88.2} &  81.5 \\
GPT-3.5       &          73.3 &             72.5 &        74.4 &          76.2 &   74.1 &      \textbf{84.3} &          90.0 &        \textbf{90.6} &          90.0 &                      84.5 &     87.9 &  81.0 \\
GPT-4         &          \textbf{92.5} &             \textbf{84.2} &        \textbf{87.5} &          \textbf{88.8} &   \textbf{88.2} &      83.0 &          86.7 &        88.8 &          \textbf{92.1} &                      85.2 &     87.2 &  \textbf{87.7} \\
\bottomrule
\end{tabular}%
\caption{Accuracy results (\%) of identifying fallacious steps on \OurDataset.}
\vspace{-3mm}
\label{tab:step_fallacy}
\end{table*}

\begin{table}[t!]
\small
\centering
\resizebox{\linewidth}{!}{%
\begin{tabular}{@{}p{8.9cm}@{}}
\toprule
\textbf{Formal $\rightarrow$ Proposition $\rightarrow$ Denying the Antecedent}  \\
\textcolor{red!90!black}{$\bullet$ Since [if you have a pinna, then you can hear] and [you do not have a pinna], therefore, [you cannot hear].} \hfill \textbf{Prediction:} Yes. \XSolidBrush \\
\textcolor{green!60!black}{$\bullet$ Since [if you do not have a pinna, then you cannot hear] and [you do not have a pinna], therefore, [you cannot hear].}  \hfill \textbf{Prediction:} Yes. \CheckmarkBold\\ 
\midrule[0.8pt]
\textbf{Formal $\rightarrow$ Syllogism $\rightarrow$ Exclusive Premises}  \\ 
\textcolor{red!90!black}{$\bullet$ Since [no psychologists are proponents of shock therapy] and [some proponents of shock therapy are not doctors], therefore, [some doctors are not psychologists].}   \hfill \textbf{Prediction:} Yes. \XSolidBrush \\
\textcolor{green!60!black}{$\bullet$ Since [no psychologists are proponents of shock therapy] and [some doctors are proponents of shock therapy], therefore, [some doctors are not psychologists].}  \hfill \textbf{Prediction:} Yes. \CheckmarkBold\\  \midrule[0.8pt]
\textbf{Informal $\rightarrow$ Irrelevance $\rightarrow$ Appeal to Pity
}  \\
\textcolor{red!90!black}{$\bullet$ Since [my horse's stirrups are broken] and [I thus had to make a pitiful 10-mile walk in the pouring rain to get home], therefore, [the broken stirrups should be replaced by the store for free].} \hfill \textbf{Prediction:} No.\CheckmarkBold\\ 
\textcolor{green!60!black}{$\bullet$ Since [my horse's stirrups are broken] and [the stirrups were under warranty], therefore, [the broken stirrups should be replaced by the store for free].}   \hfill \textbf{Prediction:} Yes. \CheckmarkBold \\ 
\midrule[0.8pt]
\textbf{Informal $\rightarrow$ Insufficiency $\rightarrow$ Questionable Cause} \\
\textcolor{red!90!black}{$\bullet$ Since [grebos are often seen during rainstorms] and [rainstorms cause floods], therefore, [grebos cause floods].} \hfill \textbf{Prediction:} No.\CheckmarkBold\\ 
\textcolor{green!60!black}{$\bullet$ Since [grebos are often seen during rainstorms] and [rainstorms cause floods],\! therefore,\! [grebos can be seen during floods].} \!\textbf{Prediction}:No. \XSolidBrush \\
\bottomrule
\end{tabular}%
}
\caption{Contrastive reasoning steps in \OurDataset and verification predictions of GPT-4.  \textcolor{red!90!black}{Fallacious steps} (the first sentence of each cell) are in red and \textcolor{green!60!black}{correct ones} (the second sentence of each cell) in green. \CheckmarkBold and \XSolidBrush indicates whether the prediction matches the label or not.}
\vspace{-3mm}
\label{tab:cases}
\end{table}

\section{Experiments}

\subsection{Models and Prompts}

We test a range of common LLMs, including GPT-4~\cite{DBLP:journals/corr/abs-2303-08774}, GPT-3.5~\cite{GPT-3.5}, Llama2~\cite{DBLP:journals/corr/abs-2307-09288}, Vicuna~\cite{DBLP:journals/corr/abs-2306-05685}, WizardLM~\cite{DBLP:journals/corr/abs-2304-12244}, Flan-T5~\cite{DBLP:journals/corr/abs-2210-11416}, Falcon~\cite{falcon40b}, Baichuan2~\cite{DBLP:journals/corr/abs-2309-10305}, ChatGLM~\cite{DBLP:conf/acl/DuQLDQY022}, and InternLM~\cite{2023internlm}.
We use the default generation parameters in the models' configuration files.\footnote{For GPT-4 and GPT-3.5, we set the temperature parameter to 0. Experiments were conducted mainly in Nov. 2023.}
Details about the models (e.g., the version numbers) can be found in Table~\ref{tab:exp_models} in the Appendix.
We evaluate the models using the same simple prompt in the zero-shot setting. 
For the prompt selection, we followed the experience of previous work~\cite{DBLP:journals/csur/LiuYFJHN23} and carefully designed and experimented with several different styles of prompts.
In the end, we chose a prompt that worked well for all models as our prompt.
Appendix~\ref{sec:appendix_exp_detail} presents a detailed ablation analysis of prompt selection, as well as experiments under few-shot setting. 

{\small
\begin{tcolorbox}[colback=gray!10,colframe=blue!70!black,
left=7pt,right=7pt,top=2pt,bottom=2pt,
float, floatplacement=t,
enlarge bottom by=-3mm,
title=Prompt for Identifying Fallacious Steps]
Is the following reasoning step correct? You can only answer "Yes" or "No." \\
\{reasoning step\}
\end{tcolorbox}
}

\subsection{Metrics}
\label{sec:data_metric}

We evaluate LLMs on 4,640 steps in \OurDataset. 
We calculate the accuracy for each type of fallacy separately. 
The accuracy of a higher-level category is the (macro) average of the accuracies of its subcategories. 
Ultimately, we take the average accuracy on formal and informal categories as the overall accuracy.

\subsection{Can LLMs accurately identify fallacious steps?}
\label{sec:exp_1}
Table~\ref{tab:step_fallacy} shows the accuracies of the different models for identifying fallacious steps on \OurDataset.
Based on the experimental results, we have the following observations.

\textbf{Identifying fallacious steps is still challenging for LLMs.}
Most LLMs struggle with accurately identifying the fallacious steps.
As shown in Table~\ref{tab:step_fallacy}, the performance of most LLMs in this binary classification task ranges from 60\% to 80\%, indicating the complexity of this task. 
The best result is achieved by GPT-4, which reaches an overall average accuracy of 87.7\%.
However, this may still fall short of guaranteeing the validity of the self-verification approach since it is only the performance of identifying single-step fallacies. 
\rw{For a long argument comprising multiple reasoning steps, the overall verification performance for the entire argument could be the product of the verification performances of each individual step. 
Consequently, the overall verification performance of the argument might decrease exponentially with the number of steps in it. }
\rw{These results suggest that we need further research on existing self-verification methods to understand how they work and under what situations they can provide correct verification feedback.}

\textbf{Formal fallacy is more difficult than informal fallacy for LLMs.}
The performance of most LLMs on formal fallacies is much lower than on informal fallacies. 
For example, GPT-3.5 achieves 87.9\% accuracy on formal fallacies, while it achieves only 74.1\% accuracy on informal fallacies.
As stated in Sec.~\ref{sec:taxonomy_of_fallacy}, formal fallacies are more related to the logical structure of reasoning and require a greater emphasis on the understanding and utilization of formal logic.
On the other hand, informal fallacies focus more on the content and semantics of the reasoning and may involve factors such as linguistic expression, semantic understanding, and semantic relevance.
Therefore, the differences in the performance of LLMs across different types of fallacies may stem from their ability to understand logical structures and semantic meanings, 
i.e., the models present some challenges in dealing with the logical structures under the natural language, whereas they perform better in dealing with issues related to content and semantics.
Taking a more fine-grained view, among formal fallacies, most models typically perform well on syllogism fallacies and poorly on proposition and probability fallacies. 
For informal fallacies, most models perform worse on ambiguity fallacies and have closer performance on the remaining four sub-categories.

\begin{table*}[t]
\small
\centering
\begin{tabular}{@{}l|XXXXY|XXXXXY|Z}
\toprule
\multirow{2}{*}{\textbf{Model}} & \multicolumn{5}{c|}{\textbf{Formal}} & \multicolumn{6}{c|}{\textbf{Informal}} & \multicolumn{1}{c}{} \\ 
 & \multicolumn{1}{c}{\textbf{prop.}} & \multicolumn{1}{c}{\textbf{quant.}} & \multicolumn{1}{c}{\textbf{syl.}} & \multicolumn{1}{c}{\textbf{prob.}} & \multicolumn{1}{c|}{\textbf{Avg.}} & \multicolumn{1}{c}{\textbf{amb.}} & \multicolumn{1}{c}{\textbf{incon.}} & \multicolumn{1}{c}{\textbf{irrel.}} & \multicolumn{1}{c}{\textbf{insuf.}} & \multicolumn{1}{c}{\textbf{inappr.}} & \multicolumn{1}{c|}{\textbf{Avg.}} & \multicolumn{1}{c}{\multirow{-2}{*}{\textbf{Avg.}}} \\ \midrule
Random &          0.4 &             0.4 &        0.4 &          0.4 &   0.4 &      0.4 &          0.4 &        0.4 &          0.4 &                      0.4 &     0.4 &  0.4 \\
\midrule
Flan-T5-Large &          15.0 &              1.7 &         0.0 &           0.0 &    4.2 &       0.0 &           0.0 &         7.7 &           4.7 &                       3.9 &      3.2 &   3.7 \\
Flan-T5-xl    &          21.7 &              0.0 &         0.0 &           5.0 &    6.7 &       0.0 &           0.0 &        16.4 &          10.7 &                       7.8 &      7.0 &   6.8 \\
Flan-T5-xxl   &          28.3 &             26.7 &         0.0 &          22.5 &   19.4 &       2.7 &          \textbf{16.7} &         8.5 &           9.5 &                       7.4 &      8.9 &  14.2 \\
Llama2-7B     &          16.7 &              0.0 &         0.0 &           0.0 &    4.2 &       0.0 &           0.0 &         1.0 &           0.0 &                       0.0 &      0.2 &   2.2 \\
Llama2-13B    &           8.3 &              0.0 &         0.0 &           0.0 &    2.1 &       0.0 &           0.0 &         0.0 &           0.0 &                       0.2 &      0.0 &   1.1 \\
Llama2-70B    &          30.0 &              0.0 &         0.0 &           0.0 &    7.5 &       4.7 &           0.0 &        14.6 &           8.1 &                      10.0 &      7.5 &   7.5 \\S
Baichuan2-7B  &           0.0 &              0.0 &         1.2 &           0.0 &    0.3 &       0.0 &           0.0 &         3.5 &           0.9 &                       3.0 &      1.5 &   0.9 \\
Baichuan2-13B &           0.0 &              0.0 &         0.0 &           2.5 &    0.6 &       0.0 &           0.0 &        13.7 &           3.1 &                       4.6 &      4.3 &   2.5 \\
ChatGLM-6B    &           0.0 &              0.0 &         0.0 &           0.0 &    0.0 &       0.0 &           0.0 &         0.0 &           0.0 &                       0.0 &      0.0 &   0.0 \\
ChatGLM2-6B   &           0.0 &              0.0 &         0.0 &           2.5 &    0.6 &       0.0 &           0.0 &         1.0 &           0.0 &                       0.0 &      0.2 &   0.4 \\
InternLM-7B   &           0.0 &              0.0 &         0.0 &           0.0 &    0.0 &       0.0 &           0.0 &         0.3 &           0.0 &                       0.4 &      0.1 &   0.1 \\
InternLM-20B  &           3.3 &              1.7 &         2.5 &          12.5 &    5.0 &       0.7 &           0.0 &         4.9 &           1.0 &                       2.2 &      1.8 &   3.4 \\
Falcon-7B     &           0.0 &              0.0 &         0.0 &           0.0 &    0.0 &       0.0 &           0.0 &         0.1 &           1.0 &                       0.0 &      0.2 &   0.1 \\
WizardLM-13B  &          15.0 &              0.0 &         0.0 &           7.5 &    5.6 &       0.0 &           0.0 &         1.0 &           0.5 &                       0.2 &      0.3 &   3.0 \\
Vicuna-7B     &           0.0 &              0.0 &         0.0 &           0.0 &    0.0 &       0.0 &           0.0 &         0.5 &           0.0 &                       0.0 &      0.1 &   0.1 \\
Vicuna-13B    &          31.7 &              0.0 &         0.0 &           0.0 &    7.9 &       0.0 &           0.0 &         2.2 &           0.0 &                       0.6 &      0.5 &   4.2 \\
Qwen-14B      &          10.0 &              0.0 &         0.0 &          20.0 &    7.5 &       0.0 &           0.0 &         5.5 &           2.1 &                       0.6 &      1.6 &   4.6 \\
GPT-3.5       &          40.0 &             20.0 &         3.8 &          22.5 &   21.6 &      16.0 &           0.0 &        22.7 &          16.4 &                       9.6 &     12.9 &  17.3 \\
GPT-4         &          \textbf{58.3} &             \textbf{31.7} &        \textbf{20.0} &          \textbf{55.0} &   \textbf{41.2} &      \textbf{40.0} &          \textbf{16.7} &        \textbf{31.9} &          \textbf{27.2} &                      \textbf{27.8} &     \textbf{28.7} &  \textbf{35.0} \\
\bottomrule
\end{tabular}%
\caption{Accuracy results (\%) on classifying the fallacy types of fallacious steps.}
\vspace{-3mm}
\label{tab:classify}
\end{table*}

\textbf{The performance of the same model on different types of fallacies can be remarkably imbalanced.}
For example, the model Qwen-14B achieves an impressive 91.7\% accuracy on the inconsistency fallacies but drops to a mere 67.5\% on probability fallacies.
This highlights a key observation that models may have superior verification abilities on some types of fallacy but are not necessarily equally well adapted to other types of fallacy.
Such imbalance performance could be particularly important for practical applications, as different types of fallacies are not always evenly distributed in a given dataset or scenario, and certain types of fallacies might be more frequent.
Therefore, we should not rely on a particular dataset when using or researching self-verification methods. Instead, we need to comprehensively consider the performance of the methods in dealing with different fallacies and scenarios.

\textbf{GPT-4 achieves superior performance, particularly in identifying formal fallacies.}
Compared to the other models, GPT-4 achieves the best results in overall average accuracy.
This gap is insignificant on informal fallacies, where models such as WizardLM-13B, Vicuna-13B, and Qwen-14B have comparable or even better average accuracies than GPT-4 on informal fallacies.
However, on formal fallacies, GPT-4's accuracy is 13.5\% higher than the second-best model  (88.2\% for GPT-4 compared to 74.7\% for the second-best model Qwen-14B).
The results suggest that GPT-4 demonstrates superior abilities in identifying fallacies related to logical structures than other LLMs.
However, there is still room to improve the performance of GPT-4.
Table~\ref{tab:cases} demonstrates the prediction results of GPT-4 in some cases.
It can be seen that GPT-4 can also fail on some challenging samples.

{ \small
\begin{tcolorbox}[colback=gray!10,colframe=blue!70!black,
left=7pt,right=7pt,top=2pt,bottom=2pt,
float, floatplacement=t,
enlarge bottom by=-3mm,
title=Prompt for Classifying Fallacy Types]
You are a logical fallacy classifier. Given an incorrect reasoning step, your task is to identify its type of fallacy. \\
Answer by choosing one of these fallacies:  \\
\{(1) Affirming the Consequent \\
(2) Denying the Antecedent \\
$\cdots$ $\cdots$ \\
(232) Alleged Certainty\} \\
You should only answer the name of the fallacy.\\
What type of fallacy does the following reasoning step belong to?\\
\{reasoning step\}
\end{tcolorbox}
}

\subsection{Can LLMs distinguish types of logical fallacies?}

In addition to whether LLMs can identify between correct and incorrect, we are also interested in whether LLMs can distinguish between different types of fallacies.
Classifying types of fallacies requires the model to understand not only the pattern of errors in reasoning but also where the errors occur and why they may occur, which requires a higher level of reasoning ability.
Given an error reasoning step, we require the model to recognize the error pattern within it and classify it as one of the 232 fallacies.
We conduct experiments on the 2,320 fallacy steps in \OurDataset and calculate the macro average accuracy.
Previous work has also explored this task and named it ``logical fallacy detection''~\cite{jin-etal-2022-logical}.
However, they only classify over 13 types of fallacies, whereas our task requires classifying over 232 types.
Our task is more challenging and allows for fine-grained and hierarchical analysis. 
We evaluate the model using the same prompt in a zero-shot setting.

\begin{table}[t!]
\centering
\resizebox{0.95\linewidth}{!}{%
\begin{tabular}{@{}l|YY|Z}
\toprule
\multicolumn{1}{@{}l|}{\textbf{Model}}  & \multicolumn{1}{c}{\textbf{Formal}}  & \multicolumn{1}{c|}{\textbf{Informal}}  & \multicolumn{1}{c}{\textbf{Avg.}}  \\ 
\midrule
Flan-T5-Large  & 62.3 (+1.3) & 65.5 (-8.1) & 63.9 (-3.4) \\
Flan-T5-xl & 56.6 (-11.3) & 67.1 (-7.0) & 61.8 (-9.2) \\
Flan-T5-xxl & 66.2 (+2.3) & 73.1 (+0.1) & 69.7 (+1.3) \\
Llama2-7B & 50.9 (-10.4) & 52.0 (-19.8) & 51.5 (-15.0) \\
Llama2-13B & 57.9 (-2.0) & 53.2 (-10.0) & 55.6 (-5.9) \\
Llama2-70B & 54.0 (-19.3) & 56.9 (-31.3) & 55.4 (-25.3) \\
Baichuan2-7B & 52.8 (+0.1) & 58.0 (+5.7) & 55.4 (+2.9) \\
Baichuan2-13B & 50.1 (-17.9) & 52.7 (-28.2) & 51.4 (-23.0) \\
ChatGLM-6B & 54.6 (-1.2) & 58.0 (+3.6) & 56.3 (+1.2) \\
ChatGLM2-6B & 58.0 (-3.7) & 61.9 (-13.3) & 60.0 (-8.4) \\
InternLM-7B & 55.1 (-5.5) & 59.4 (-16.0) & 57.2 (-10.8) \\
InternLM-20B & 59.2 (-4.4) & 68.6 (-8.9) & 63.9 (-6.7) \\
Falcon-7B & 40.2 (-2.8) & 46.9 (-23.3) & 43.5 (-13.1) \\
WizardLM-13B & 74.2 (+6.7) & 82.9 (-3.6) & 78.5 (+1.5) \\
Vicuna-7B & 70.4 (+2.0) & 78.0 (-4.2) & 74.2 (-1.1) \\
Vicuna-13B & 61.6 (-8.2) & 67.7 (-18.7) & 64.6 (-13.5) \\
Qwen-14B & 71.1 (-3.6) & 79.2 (-9.0) & 75.2 (-6.3) \\
GPT-3.5 & 75.1 (+1.0) & 75.7 (-12.2) & 75.4 (-5.6) \\
GPT-4 & 91.2 (+3.0) & 84.9 (-2.3) & 88.0 (+0.3) \\
 \bottomrule
\end{tabular}%
}
\caption{Accuracy results (\%) of identifying fallacious steps given the definitions of corresponding fallacies. We present the performance variations in the case with the definitions in parentheses compared to the case without definitions.}
\label{tab:give_definition}
\end{table}

Table~\ref{tab:classify} shows the results. 
First, this task is very challenging for the existing LLMs. 
The models' performances are poor, with less than 10\% overall accuracy, except for three models, GPT-4, GPT-3.5, and Flan-T5-xxl.
For example, Vicuna-13B can achieve 78.1\% accuracy on identifying fallacious steps, but only 4.2\% accuracy on this task.
Among all the models, GPT-4 performs the best, achieving an overall accuracy of 35.0\%. 
This indicates that GPT-4 can recognize and classify the reasoning error patterns to a certain degree, showing a stronger reasoning ability than other models.
Nevertheless, there is still substantial room for improvement. 
Further research may be required to achieve higher accuracy and enhance reasoning ability.

It is worth noting that on this task, the models typically perform better on formal than informal fallacies. 
This is inconsistent with the observation in identifying fallacious steps (Table~\ref{tab:step_fallacy}).
When identifying fallacious steps, models typically perform worse on formal fallacies and better on informal fallacies.
There could be various reasons for this inconsistency.
One possible explanation is that the models might just know the names of the fallacies rather than having an in-depth understanding of what these fallacies are.
In determining whether there is an error in reasoning, the models might not be relying directly on their understanding of reasoning or fallacies but on some other abilities, which thus contributes to this inconsistency.

\subsection{Can LLMs understand fallacies better from their definitions?}

We test whether LLMs can perform better in identifying fallacious steps given the fallacy definition.
Specifically, for each step in \OurDataset, we add the name and definition of its corresponding fallacy to the prompt of LLMs in advance.
We then ask the model to determine whether the reasoning step is correct as before.
As stated in Sec.~\ref{sec:data_collection}, the definitions of fallacies are gathered from academic sources.
An example of the definition can be found in Figure~\ref{fig:fallacy_definition} in the Appendix.

{\small
\begin{tcolorbox}[colback=gray!10,colframe=blue!70!black,
left=7pt,right=7pt,top=2pt,bottom=2pt,
float, floatplacement=t,
title=Prompt for Identifying Fallacious Steps Given Fallacy Definition]
You are a trained model capable of identifying the logical fallacy known as \{fallacy\}.\\
This is the definition for \{fallacy\}: \{fallacy definition\}\\
Is the following reasoning step correct? You can only answer "Yes" or "No."\\
\{reasoning step\}
\end{tcolorbox}
}

We can observe a surprising trend by analyzing the results in Table~\ref{tab:give_definition}.
When definitions of corresponding fallacies are provided in advance, most models' performance decreases rather than improves.
For instance, the overall accuracy of Vicuna-13B decreases from 78.1\% to 64.6\%, with a 13.5\% decrease.
These results suggest that providing definitions may hurt models' performance.

The reasons for this phenomenon deserve further exploration.
One possible reason is that, in pre-trained data, the definitions and the fallacies themselves may not co-occur frequently, resulting in a mismatch with the current setting.
Moreover, the mechanism by which the models judge fallacies has not yet been fully clarified.
In this case, even if the definitions are provided, the models fail to improve performance.
Instead, the prompt becomes complex with the addition of definitions, possibly interfering with their decision-making process.
These observations inspire us that more intensive research is called to understand what are the mechanisms by which LLMs understand the reasoning and fallacies.

\section{Conclusion}
In this paper, we take a closer look at the verification abilities of LLMs in logical reasoning.
We collect a dataset containing 232 fallacies and propose a hierarchical taxonomy of fallacies.
Our main experimental finding is that most LLMs still struggle to identify fallacies in logical reasoning accurately.
This implies that it may be overly optimistic to expect LLMs to be able to inherently identify errors and conduct self-verification reasoning, at least with respect to the current state of technology.
Therefore, researchers and practitioners should be more cautious in using self-verification methods.
We call for more research to explore the potentials and limitations of self-verification methods to steer LLMs towards improved accuracy and reliability.

\section*{Acknowledgements}
\rw{We appreciate the anonymous reviewers for their insightful comments.}
\rw{This work is supported by the Natural Science Foundation of China (NSFC No. 62176132) and the Guoqiang Institute of Tsinghua University (2020GQG0005).}

\section*{Limitations}

In this paper, we present an intensive investigation and evaluation of the verification abilities of Large Language Models (LLMs) for logical reasoning.
Although we have reached some findings, we are also aware that there is still some room for improvement and future research areas worth exploring.

Firstly, we conduct experiments only on the most common LLMs.
Such a limitation comes from two main reasons: one is due to the limitation of computational resources of our research team; and the other is the barrier of access to certain closed-source models.
This results in our inability to perform detailed experiments on all types and all scales of models.
Thus, our results may not fully reflect the abilities of all LLMs.
In future studies, it is worthwhile to turn our attention to more types and scales of models to provide a more comprehensive evaluation of their abilities.

Second, our study focused mainly on the aspect of logical reasoning.
Reasoning in real-world applications often encompasses other types of reasoning, such as numerical reasoning.
It would be interesting to extend our research to more types of reasoning.
By doing so, we can reveal the boundaries of the abilities of LLMs in these areas, which can enhance our deeper understanding of the performance of LLMs.
Meanwhile, it would also help us to figure out how we can improve the accuracy and robustness of the reasoning of LLMs.

This article follows the ACL Code of Ethics. To the best of our knowledge, our work is foundational research, and we do not find obvious risks related to malicious harmful effects, environmental impact, fairness considerations, or privacy considerations.

\bibliography{anthology,custom}
\bibliographystyle{acl_natbib}

\newpage
\clearpage

\appendix

\section{Details of Data Collection}
\label{sec:appendix_data}

\begin{figure*}[t]
  \centering
  \includegraphics[width=\textwidth]{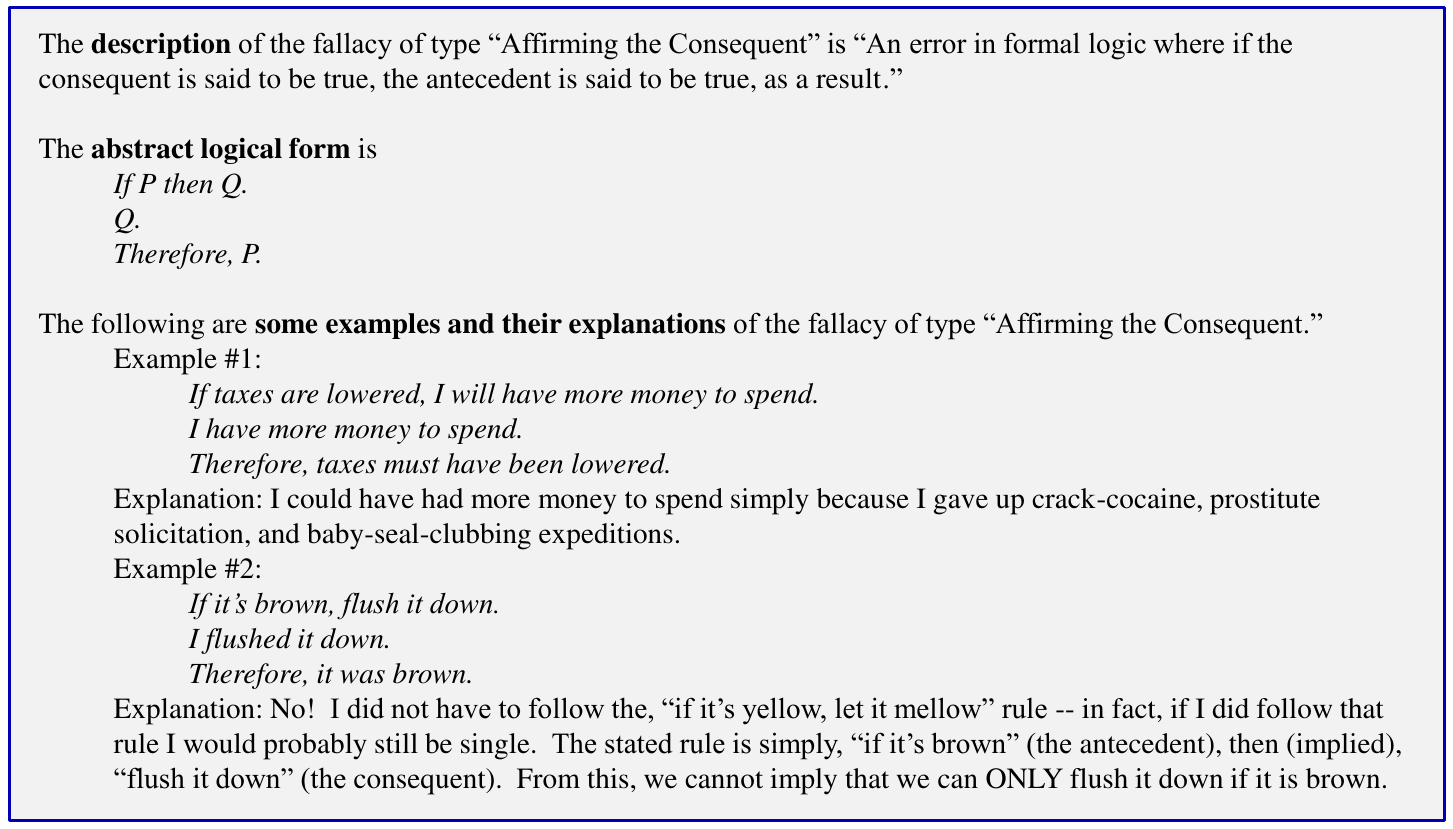}
  \caption{The definition of the fallacy of ``Affirming the Consequent'', one of 232 types of fallacies in our dataset.}
  \label{fig:fallacy_definition}
\end{figure*}

To generate candidate fallacious steps, we prompt GPT-4 with the following content.
Figure~\ref{fig:fallacy_definition} shows an example of the fallacy definition.
Note that the steps generated by GPT-4 are only used as candidates.
We then have human experts proofread and revise them to ensure the quality of the data.
\rw{We invited 10 well-trained graduate students from universities as human expert annotators. They have passed the graduate school entrance exam and have well logical reasoning skills. They have undergone rigorous training in areas including Mathematical Logic, Computer Science, Programming, and Statistics.}

{ \small
\begin{tcolorbox}[colback=gray!10,colframe=blue!70!black,
left=7pt,right=7pt,top=2pt,bottom=2pt,
title=Prompt for Generating Candidate Fallacious Steps]
You are a faulty reasoner. I will describe a logical fallacy to you, and then you generate a reasoning step that belongs to this logical fallacy.\\
Here is the description: \{fallacy definition\} \\
Now, generate a reasoning step that contains this type of logical fallacy. The generated step should be related to this element: \{entity\} \\
Use square brackets [] to include propositions. The format of the output is "Since [XXX] and [XXX], therefore, [XXX] 
\end{tcolorbox}
}

In the end of the appendix, Table~\ref{tab:fallacy_232} shows the detailed categorizations and descriptions of the 232 fallacies in our dataset.

\section{Details of Experiments}
\label{sec:appendix_exp_detail}

\subsection{Models}
Table~\ref{tab:exp_models} shows the version and source URL of the LLMs used in our experiments. 
For all series of LLMs, we select their versions of instruction fine-tuned or chat fine-tuned, since these models are closest to realistic applications.
We follow the licences of these models to use them.

\begin{table*}[t]
\centering
\resizebox{\textwidth}{!}{%
\begin{tabular}{lll}
\toprule
\textbf{Model}   & \textbf{Version}                         & \textbf{URL}                                                      \\
\midrule
Flan-T5-Large~\cite{DBLP:journals/corr/abs-2210-11416} & google/flan-t5-large            & \href{https://huggingface.co/google/flan-t5-large}{https://huggingface.co/google/flan-t5-large}            \\
Flan-T5-xl~\cite{DBLP:journals/corr/abs-2210-11416}    & google/flan-t5-xl               & \href{https://huggingface.co/google/flan-t5-xl}{https://huggingface.co/google/flan-t5-xl}               \\
Flan-T5-xxl~\cite{DBLP:journals/corr/abs-2210-11416}   & google/flan-t5-xxl              & \href{https://huggingface.co/google/flan-t5-xxl}{https://huggingface.co/google/flan-t5-xxl}              \\
\midrule
Llama2-7B~\cite{DBLP:journals/corr/abs-2307-09288}     & meta-llama/Llama-2-7b-chat-hf   & \href{https://huggingface.co/meta-llama/Llama-2-7b-chat-hf}{https://huggingface.co/meta-llama/Llama-2-7b-chat-hf}   \\
Llama2-13B~\cite{DBLP:journals/corr/abs-2307-09288}    & meta-llama/Llama-2-13b-chat-hf  & \href{https://huggingface.co/meta-llama/Llama-2-13b-chat-hf}{https://huggingface.co/meta-llama/Llama-2-13b-chat-hf}  \\
Llama2-70B~\cite{DBLP:journals/corr/abs-2307-09288}    & meta-llama/Llama-2-70b-chat-hf  & \href{https://huggingface.co/meta-llama/Llama-2-70b-chat-hf}{https://huggingface.co/meta-llama/Llama-2-70b-chat-hf}  \\
\midrule
Baichuan2-7B~\cite{DBLP:journals/corr/abs-2309-10305}   & baichuan-inc/Baichuan2-7B-Chat  & \href{https://huggingface.co/baichuan-inc/Baichuan2-7B-Chat}{https://huggingface.co/baichuan-inc/Baichuan2-7B-Chat}  \\
Baichuan2-13B~\cite{DBLP:journals/corr/abs-2309-10305}  & baichuan-inc/Baichuan2-13B-Chat & \href{https://huggingface.co/baichuan-inc/Baichuan2-13B-Chat}{https://huggingface.co/baichuan-inc/Baichuan2-13B-Chat} \\
\midrule
ChatGLM-6B~\cite{DBLP:conf/acl/DuQLDQY022}    & THUDM/chatglm-6b                & \href{https://huggingface.co/THUDM/chatglm-6b}{https://huggingface.co/THUDM/chatglm-6b}                \\
ChatGLM2-6B~\cite{DBLP:conf/acl/DuQLDQY022}   & THUDM/chatglm2-6b               & \href{https://huggingface.co/THUDM/chatglm2-6b}{https://huggingface.co/THUDM/chatglm2-6b}               \\
\midrule
InternLM-7B~\cite{2023internlm}   & internlm/internlm-chat-7b-v1\_1 & \href{https://huggingface.co/internlm/internlm-chat-7b-v1\_1}{https://huggingface.co/internlm/internlm-chat-7b-v1\_1} \\
InternLM-20B~\cite{2023internlm}  & internlm/internlm-chat-20b      & \href{https://huggingface.co/internlm/internlm-chat-20b}{https://huggingface.co/internlm/internlm-chat-20b}     \\
\midrule
Falcon-7B~\cite{falcon40b}     & tiiuae/falcon-7b-instruct       & \href{https://huggingface.co/tiiuae/falcon-7b}{https://huggingface.co/tiiuae/falcon-7b}                \\
\midrule
WizardLM-13B~\cite{DBLP:journals/corr/abs-2304-12244}  & WizardLM/WizardLM-13B-V1.2      & \href{https://huggingface.co/WizardLM/WizardLM-13B-V1.2}{https://huggingface.co/WizardLM/WizardLM-13B-V1.2}      \\
\midrule
Vicuna-7B~\cite{DBLP:journals/corr/abs-2306-05685}     & lmsys/vicuna-7b-v1.5            & \href{https://huggingface.co/lmsys/vicuna-7b-v1.5}{https://huggingface.co/lmsys/vicuna-7b-v1.5}            \\
Vicuna-13B~\cite{DBLP:journals/corr/abs-2306-05685}    & lmsys/vicuna-13b-v1.5           & \href{https://huggingface.co/lmsys/vicuna-13b-v1.5}{https://huggingface.co/lmsys/vicuna-13b-v1.5}           \\
\midrule
Qwen-14B~\cite{DBLP:journals/corr/abs-2309-16609}      & Qwen/Qwen-14B-Chat              & \href{https://huggingface.co/Qwen/Qwen-14B-Chat}{https://huggingface.co/Qwen/Qwen-14B-Chat}             \\
\midrule
GPT-3.5~\cite{GPT-3.5}       & gpt-3.5-turbo                   & \href{https://platform.openai.com/docs/models/gpt-3-5}{https://platform.openai.com/docs/models/gpt-3-5}        \\
GPT-4~\cite{DBLP:journals/corr/abs-2303-08774}         & gpt-4                           & \href{https://platform.openai.com/docs/models/gpt-4}{https://platform.openai.com/docs/models/gpt-4}   \\    
\bottomrule
\end{tabular}
}
\caption{Detailed information about the models we experiment with.}
\vspace{+2cm}
\label{tab:exp_models}
\end{table*}

\subsection{Prompt Selection}
To select suitable prompts and to explore the impact of prompts on model performance, we conduct an ablation study on the prompts.
Table~\ref{tab:prompts} demonstrates the prompts we used.
Prompt 1 is the simplest prompt. 
The model is expected to answer ``Yes'' for the correct reasoning steps and ``No'' for the incorrect reasoning steps.
Prompt 2, alternatively, replaces the response ``Yes/No'' with ``True/False.''
Prompt 3 adopts a Chain-of-Thought-style prompt~\cite{DBLP:conf/nips/KojimaGRMI22} that allows the model to generate some relevant rationales before giving the prediction.
Prompt 4 describes the task in more detail. Moreover, in contrast to Prompt 1, Prompt 4 requires the model to answer ``Yes'' for incorrect reasoning steps (which contain logical fallacies) and ``No'' to correct reasoning steps (which do not contain logical fallacies).

Table~\ref{tab:prompt_ablation} demonstrates the results.
We can find that Prompt 1 works well for all models.
For Prompt 2, some of the models show significant performance degradation after replacing the response words.
Prompt 3 introduces chains of thought, but does not achieve significant performance gains on all models.
For Prompt 4, although it describes the task in more detail, it does not seem to help improve the LLMs' performance.
Among all the models, GPT-4 exhibits the strongest robustness to prompts, achieving similar performance with different prompts.
Comprehensively, we finally chose Prompt 1 as our prompt.

\subsection{Few-shot Setting}
We also conduct experiments under few-shot setting.
We include four demonstrations in the prompt, covering the correct and fallacious steps related to formal and informal fallacies.

{ \small
\begin{tcolorbox}[colback=gray!10,colframe=blue!70!black,
left=7pt,right=7pt,top=2pt,bottom=2pt,
title=Prompt for Identifying Fallacious Steps under Few-shot Setting]
Is the following reasoning step correct? You can only answer "Yes" or "No." \\
Since [If it's raining then the streets are wet] and [It's raining now], therefore, [The streets are wet]. \\ Yes. \\
Since [I found a shell on the beach] and [this shell was beautifully shaped and colored], therefore, [all shells are beautifully shaped and colored]. \\ No. \\
Since [I am at home or I am in the city] and [I am at home], therefore, [I am not in the city]. \\ No. \\
Since [heavy snowfall often leads to traffic jams] and [traffic jams cause delays], therefore, [heavy snowfall can lead to delays]. \\ Yes. \\
\{reasoning step\}
\end{tcolorbox}
}

Table~\ref{tab:step_fallacy_few_shot} demonstrates the performance of identifying fallacious steps under few-shot setting.
We can find that most of the LLMs achieve comparable performance under few-shot setting compared to that under zero-shot setting. 
Moreover, our findings in Sec~\ref{sec:exp_1} stand under the few-shot setting as well.

\begin{table*}[t]
\small
\centering
\begin{tabular}{@{}lp{13cm}}
\toprule
\textbf{No.} & \textbf{Content} \\ \midrule
\textbf{Prompt 1} & Is the following reasoning step correct? \textbackslash n You can only answer "Yes" or "No".\textbackslash n \{reasoning step\} \\ \midrule
\textbf{Prompt 2} & Is the following reasoning step correct? \textbackslash n You can only answer "True" or "False".\textbackslash n \{reasoning step\} \\ \midrule
\textbf{Prompt 3} & Is the following reasoning step correct? \textbackslash n Let's think step by step and then answer "Yes" or "No".\textbackslash n \{reasoning step\} \\ \midrule
\textbf{Prompt 4} & You are a trained model capable of detecting reasoning errors and logical fallacies. \textbackslash n As a detector, your task is to analyze the given reasoning steps and determine whether they involve any logical fallacies. \textbackslash n If a logical fallacy is present, your response should be "Yes". \textbackslash n If no logical fallacies are detected, your response should be "No".\textbackslash n You can only answer "Yes" or "No". \textbackslash n \{reasoning step\} \\ 
\bottomrule
\end{tabular}%
\caption{The different prompts used to prompt large language models to identify fallacious steps.}
\label{tab:prompts}
\end{table*}

\begin{table*}[t]
\small
\centering
\resizebox{\textwidth}{!}{%
\begin{tabular}{@{}l|YYZ|YYZ|YYZ|YYZ}
\toprule
\multirow{2}{*}{\textbf{Model}} & \multicolumn{3}{c|}{\textbf{Prompt 1}} & \multicolumn{3}{c|}{\textbf{Prompt 2}} & \multicolumn{3}{c|}{\textbf{Prompt 3}} & \multicolumn{3}{c}{\textbf{Prompt 4}} \\
 & \multicolumn{1}{c}{\textbf{Formal}} & \multicolumn{1}{c}{\textbf{Informal}} & \multicolumn{1}{c|}{\textbf{Avg.}} & \multicolumn{1}{c}{\textbf{Formal}} & \multicolumn{1}{c}{\textbf{Informal}} & \multicolumn{1}{c|}{\textbf{Avg.}} & \multicolumn{1}{c}{\textbf{Formal}} & \multicolumn{1}{c}{\textbf{Informal}} & \multicolumn{1}{c|}{\textbf{Avg.}} & \multicolumn{1}{c}{\textbf{Formal}} & \multicolumn{1}{c}{\textbf{Informal}} & \multicolumn{1}{c}{\textbf{Avg.}}\\ \midrule
Flan-T5-Large & 61.0 & 73.6 & 67.3 & 59.5 & 71.7 & 65.6 & 60.9 & 73.9 & 67.4 & 40.1 & 41.3 & 40.7 \\
Flan-T5-xl & 67.9 & 74.1 & 71.0 & 68.2 & 68.5 & 68.4 & 67.3 & 74.7 & 71.0 & 41.0 & 47.6 & 44.3 \\
Flan-T5-xxl & 63.9 & 73.0 & 68.4 & 65.1 & 74.2 & 69.6 & 63.4 & 73.6 & 68.5 & 43.1 & 59.1 & 51.1 \\
Llama2-7B & 61.3 & 71.8 & 66.5 & 50.7 & 53.3 & 52.0 & 49.8 & 52.2 & 51.0 & 50.0 & 49.9 & 50.0 \\
Llama2-13B & 59.9 & 63.2 & 61.5 & 57.5 & 67.3 & 62.4 & 49.8 & 51.0 & 50.4 & 50.2 & 51.1 & 50.7 \\
Llama2-70B & 73.3 & 88.2 & 80.7 & 72.3 & 87.8 & 80.1 & 63.8 & 67.7 & 65.7 & 54.7 & 59.1 & 56.9  \\
Baichuan2-7B & 52.7 & 52.3 & 52.5 & 60.6 & 75.0 & 67.8 & 54.4 & 71.1 & 62.7 & 45.3 & 46.5 & 45.9 \\
Baichuan2-13B & 68.0 & 80.9 & 74.4 & 61.7 & 82.2 & 72.0 & 65.9 & 75.9 & 70.9 & 50.9 & 49.9 & 50.4 \\
ChatGLM-6B & 55.8 & 54.4 & 55.1 & 42.9 & 48.1 & 45.5 & 53.3 & 61.3 & 57.3 & 48.2 & 46.3 & 47.3 \\
ChatGLM2-6B & 61.7 & 75.2 & 68.4 & 59.1 & 76.5 & 67.8 & 61.8 & 71.3 & 66.5 & 50.5 & 50.5 & 50.5 \\
InternLM-7B & 60.6 & 75.4 & 68.0 & 57.6 & 72.6 & 65.1 & 59.5 & 75.4 & 67.4 & 56.7 & 68.3 & 62.5 \\
InternLM-20B & 63.6 & 77.5 & 70.6 & 68.4 & 81.3 & 74.8 & 64.6 & 77.8 & 71.2 & 59.6 & 77.1 & 68.3 \\
Falcon-7B & 43.0 & 70.2 & 56.6 & 0.0 & 0.0 & 0.0 & 56.5 & 73.1 & 64.8 & 44.4 & 43.3 & 43.9 \\
WizardLM-13B & 67.5 & 86.5 & 77.0 & 44.1 & 73.5 & 58.8 & 68.5 & 85.9 & 77.2 & 56.6 & 76.2 & 66.4 \\
Vicuna-7B & 68.4 & 82.2 & 75.3 & 0.0 & 1.0 & 0.5 & 66.0 & 81.4 & 73.7 & 55.8 & 57.3 & 56.5 \\
Vicuna-13B & 69.8 & 86.4 & 78.1 & 74.2 & 85.0 & 79.6 & 60.6 & 81.4 & 71.0 & 55.7 & 67.6 & 61.7 \\
Qwen-14B & 74.7 & 88.2 & 81.5 & 64.1 & 67.9 & 66.0 & 60.2 & 79.3 & 69.7 & 57.8 & 72.3 & 65.1 \\
GPT-3.5 & 74.1 & 87.9 & 81.0 & 80.9 & 88.2 & 84.6 & 78.6 & 88.8 & 83.7 & 53.1 & 57.7 & 55.4 \\
GPT-4 & 88.2 & 87.2 & 87.7 & 89.8 & 87.4 & 88.6 & 89.0 & 87.0 & 88.0 & 82.3 & 90.8 & 86.6 \\
\bottomrule
\end{tabular}%
}
\caption{Accuracy results (\%) on identifying fallacious steps using different prompts.}
\label{tab:prompt_ablation}
\end{table*}
\begin{table*}[t!]
\small
\centering
\begin{tabular}{@{}l|XXXXY|XXXXXY|Z}
\toprule
\multirow{2}{*}{\textbf{Model}} & \multicolumn{5}{c|}{\textbf{Formal}} & \multicolumn{6}{c|}{\textbf{Informal}} & \multicolumn{1}{c}{} \\ 
 & \multicolumn{1}{c}{\textbf{prop.}} & \multicolumn{1}{c}{\textbf{quant.}} & \multicolumn{1}{c}{\textbf{syl.}} & \multicolumn{1}{c}{\textbf{prob.}} & \multicolumn{1}{c|}{\textbf{Avg.}} & \multicolumn{1}{c}{\textbf{amb.}} & \multicolumn{1}{c}{\textbf{incon.}} & \multicolumn{1}{c}{\textbf{irrel.}} & \multicolumn{1}{c}{\textbf{insuf.}} & \multicolumn{1}{c}{\textbf{inappr.}} & \multicolumn{1}{c|}{\textbf{Avg.}} & \multicolumn{1}{c}{\multirow{-2}{*}{\textbf{Avg.}}} \\ \midrule
Random &          50.0 &             50.0 &        50.0 &          50.0 &   50.0 &      50.0 &          50.0 &        50.0 &          50.0 &                      50.0 &     50.0 &  50.0 \\
\midrule
Flan-T5-Large &          58.3 &             64.2 &        71.9 &          62.5 &   64.2 &      62.3 &          85.0 &        73.8 &          73.3 &                      69.7 &     72.8 &  68.5 \\
Flan-T5-xl    &          47.5 &             74.2 &        81.9 &          68.8 &   68.1 &      65.3 &          85.0 &        70.0 &          70.3 &                      70.6 &     72.3 &  70.2 \\
Flan-T5-xxl   &          44.2 &             70.8 &        79.4 &          56.2 &   62.7 &      64.0 &          76.7 &        73.5 &          74.6 &                      71.4 &     72.0 &  67.3 \\
Llama2-7B     &          65.0 &             66.7 &        56.2 &          63.8 &   62.9 &      80.3 &          75.0 &        81.5 &          87.0 &                      81.8 &     81.1 &  72.0 \\
Llama2-13B    &          65.8 &             65.8 &        60.0 &          62.5 &   63.5 &      69.0 &          73.3 &        78.6 &          80.3 &                      80.6 &     76.4 &  70.0 \\
Llama2-70B    &          64.2 &             77.5 &        78.8 &          72.5 &   73.2 &      83.3 &          81.7 &        86.8 &          90.9 &                      88.1 &     86.2 &  79.7 \\
Baichuan2-7B  &          55.0 &             70.0 &        62.5 &          62.5 &   62.5 &      73.3 &          76.7 &        72.8 &          77.4 &                      73.9 &     74.8 &  68.7 \\
Baichuan2-13B &          60.8 &             72.5 &        68.8 &          62.5 &   66.1 &      76.3 &          75.0 &        78.7 &          82.3 &                      79.1 &     78.3 &  72.2 \\
ChatGLM-6B    &          71.7 &             72.5 &        68.1 &          61.3 &   68.4 &      70.7 &          81.7 &        73.7 &          78.5 &                      76.9 &     76.3 &  72.3 \\
ChatGLM2-6B   &          64.2 &             65.8 &        78.8 &          57.5 &   66.6 &      74.0 &          83.3 &        79.7 &          81.6 &                      76.2 &     79.0 &  72.8 \\
InternLM-7B   &          60.0 &             68.3 &        76.9 &          47.5 &   63.2 &      69.3 &          80.0 &        74.1 &          75.6 &                      71.8 &     74.2 &  68.7 \\
InternLM-20B  &          58.3 &             69.2 &        78.8 &          60.0 &   66.6 &      69.3 &          80.0 &        74.9 &          77.6 &                      75.8 &     75.5 &  71.1 \\
Falcon-7B     &          56.7 &             58.3 &        55.0 &          53.8 &   55.9 &      70.0 &          71.7 &        74.0 &          73.0 &                      72.5 &     72.2 &  64.1 \\
WizardLM-13B  &          76.7 &             72.5 &        77.5 &          71.2 &   74.5 &      82.3 &          80.0 &        85.8 &          90.8 &                      84.4 &     84.7 &  79.6 \\
Vicuna-7B     &          59.2 &             72.5 &        73.1 &          60.0 &   66.2 &      76.7 &          81.7 &        84.2 &          87.2 &                      83.5 &     82.6 &  74.4 \\
Vicuna-13B    &          67.5 &             75.0 &        83.1 &          62.5 &   72.0 &      82.7 &          85.0 &        85.8 &          90.8 &                      85.5 &     85.9 &  79.0 \\
Qwen-14B      &          68.3 &             75.0 &        83.1 &          67.5 &   73.5 &      81.7 &          88.3 &        87.8 &          91.2 &                      85.2 &     86.8 &  80.2 \\
GPT-3.5       &          72.5 &             81.7 &        78.1 &          78.8 &   77.8 &      86.0 &          86.7 &        87.9 &          89.4 &                      85.6 &     87.1 &  82.4 \\
GPT-4         &          91.7 &             84.2 &        87.5 &          88.8 &   88.0 &      84.7 &          85.0 &        89.2 &          93.0 &                      87.8 &     87.9 &  88.0 \\
\bottomrule
\end{tabular}%
\caption{Accuracy results (\%) of identifying fallacious steps under few-shot setting.}
\label{tab:step_fallacy_few_shot}
\end{table*}

\clearpage
\onecolumn

{
\renewcommand{\arraystretch}{1.5}
\small

\begin{center}
\begin{longtable}{|p{2cm}|p{3.5cm}|p{8.5cm}|}

\hline
\textbf{Category} & \textbf{Name of Fallacy}   & \textbf{Description of Fallacy} \\
\hline
\multirow[t]{6}{2cm}{Formal-Proposition} & Affirming the Consequent & An error in formal logic where if the consequent is said to be true, the antecedent is said to be true, as a result.  \\
\cline{2-3}
        & Denying the Antecedent & It is a fallacy where in a standard if/then premise, the antecedent (what comes after the “if”) is made not true, then it is concluded that the consequent (what comes after the “then”) is not true.  \\
\cline{2-3}
        & Negating Antecedent and Consequent & The valid form of this argument is as follows: If P then Q. Therefore, if not-Q then not-P.   \\
\cline{2-3}
        & Commutation of Conditionals & Switching the antecedent and the consequent in a logical argument.  \\
\cline{2-3}
        & Affirming a Disjunct & Making the false assumption that when presented with an either/or possibility, that if one of the options is true that the other one must be false.  \\
\cline{2-3}
        & Denying a Conjunct & A formal fallacy in which the first premise states that at least one of the two conjuncts (antecedent and consequent) is false and concludes that the other conjunct must be true.  \\
\hline
\multirow[t]{6}{2cm}{Formal-Quantification} & False Conversion & The formal fallacy where the subject and the predicate terms of the proposition are switched (conversion) in the conclusion, in a proposition that uses “all” in its premise (type “A” forms), or “some/not” (type “O” forms).  \\
\cline{2-3}
        & Unwarranted Contrast & Assuming that implicature means implication, when it logically does not.  \\
\cline{2-3}
        & Quantifier Shift Fallacy & A fallacy of reversing the order of two quantifiers.  \\
\cline{2-3}
        & Existential Fallacy & A formal logical fallacy, which is committed when a categorical syllogism employs two universal premises (“all”) to arrive at a particular (“some”) conclusion.  \\
\cline{2-3}
        & Fallacy of Every and All & When an argument contains both universal quantifiers and existential quantifiers (all, some, none, every) with different meanings, and the order of the quantifiers is reversed.   \\
\cline{2-3}
        & Illicit Contraposition & A formal fallacy where switching the subject and predicate terms of a categorical proposition, then negating each, results in an invalid argument form.  \\
\hline
\multirow[t]{8}{2cm}{Formal-Syllogism} & Fallacy of the Undistributed Middle & A formal fallacy in a categorical syllogism where the middle term, or the term that does not appear in the conclusion, is not distributed to the other two terms.  \\
\cline{2-3}
        & Exclusive Premises & A standard form categorical syllogism that has two negative premises either in the form of  “no X are Y” or “some X are not Y”.  \\
\cline{2-3}
        & Fallacy of Four Terms & This fallacy occurs in a categorical syllogism when the syllogism has four terms rather than the requisite three.  \\
\cline{2-3}
        & Illicit Substitution of Identicals & A fallacy due to confusing the knowing of a thing (extension) with the knowing of it under all its various names or descriptions (intension).  \\
\cline{2-3}
        & Illicit Minor & Any form of a categorical syllogism in which the minor term is distributed in the conclusion, but not in the minor premise.  \\
\cline{2-3}
        & Illicit Major & Any form of a categorical syllogism in which the major term is distributed in the conclusion, but not in the major premise.  \\
\cline{2-3}
        & Negative Conclusion from Affirmative Premises & The conclusion of a standard form categorical syllogism is negative, but both of the premises are positive.  \\
\cline{2-3}
        & Affirmative Conclusion from a Negative Premise & The conclusion of a standard form categorical syllogism is affirmative, but at least one of the premises is negative.   \\
\hline
\multirow[t]{4}{2cm}{Formal-Probability} & Gamblers Fallacy & Reasoning that, in a situation that is pure random chance, the outcome can be affected by previous outcomes.  \\
\cline{2-3}
        & Hot Hand Fallacy & The irrational belief that if you win or lose several chance games in a row, you are either “hot” or “cold,” respectively, meaning that the streak is likely to continue and has to do with something other than pure probability.  \\
\cline{2-3}
        & Conjunction Fallacy & This occurs when one estimates a conjunctive statement (this and that) to be more probable than at least one of its component statements.  \\
\cline{2-3}
        & Disjunction Fallacy & It occurs when one estimates a disjunctive statement (this or that) to be less probable than at least one of its component statements.  \\
\hline
\multirow[t]{15}{2cm}{Informal-Ambiguity} & Argument of the Beard & When one argues that no useful distinction can be made between two extremes, just because there is no definable moment or point on the spectrum where the two extremes meet.  \\
\cline{2-3}
        & Appeal to Extremes & Erroneously attempting to make a reasonable argument into an absurd one, by taking the argument to the extremes.  \\
\cline{2-3}
        & Type Token Fallacy & The type-token fallacy is committed when a word can refer to either a type (representing an abstract descriptive concept) or a token (representing an object that instantiates a concept) and is used in a way that makes it unclear which it refers to.  \\
\cline{2-3}
        & Use Mention Error & Confusing the word used to describe a thing, with the thing itself.  \\
\cline{2-3}
        & Reification & When an abstraction is treated as if it were a concrete, real event or physical entity -- when an idea is treated as if had a real existence.  \\
\cline{2-3}
        & Fake Precision & Using implausibly precise statistics to give the appearance of truth and certainty, or using a negligible difference in data to draw incorrect inferences.  \\
\cline{2-3}
        & No True Scotsman & When a universal (“all”, “every”, etc.) claim is refuted, rather than conceding the point or meaningfully revising the claim, the claim is altered by going from universal to specific, and failing to give any objective criteria for the specificity.  \\
\cline{2-3}
        & Contextomy & Removing a passage from its surrounding matter in such a way as to distort its intended meaning.  \\
\cline{2-3}
        & Stolen Concept Fallacy & Requiring the truth of the something that you are simultaneously trying to disprove.  \\
\cline{2-3}
        & Anthropomorphism & The attributing of human characteristics and purposes to inanimate objects, animals, plants, or other natural phenomena, or to gods.  \\
\cline{2-3}
        & Accent Fallacy & When the meaning of a word, sentence, or entire idea is interpreted differently by changing where the accent falls.  \\
\cline{2-3}
        & Ambiguity Fallacy & When an unclear phrase with multiple definitions is used within the argument; therefore, does not support the conclusion.  \\
\cline{2-3}
        & Alphabet Soup & The deliberate and excessive use of acronyms and abbreviations to appear more knowledgeable in the subject or confuse others.  \\
\cline{2-3}
        & Equivocation & Using an ambiguous term in more than one sense, thus making an argument misleading.  \\
\cline{2-3}
        & Modal Scope Fallacy & Making a formal argument invalid by confusing the scope of what is actually necessary or possible.  \\
\hline
\multirow[t]{3}{2cm}{Informal-Inconsistency} & Inconsistency & In terms of a fallacious argument, two or more propositions are asserted that cannot both possibly be true.  \\
\cline{2-3}
        & Conflicting Conditions & When the argument is self-contradictory and cannot possibly be true.  \\
\cline{2-3}
        & Kettle Logic & Making (usually) multiple, contradicting arguments, in an attempt to support a single point or idea.  \\
\hline
\multirow[t]{78}{2cm}{Informal-Irrelevance} & Political Correctness Fallacy & It is the assumption or admission that two or more groups, individuals, or ideas of groups or individuals, are equal, of equal value, or both true, based on the recent phenomenon of political correctness.  \\
\cline{2-3}
        & Appeal to Complexity & Concluding that because you don't understand something, it must not be true, it's improbable, or the argument must be flawed.  \\
\cline{2-3}
        & Statement of Conversion & Accepting the truth of a claim based on a conversion story without considering any evidence for the truth of the claim.  \\
\cline{2-3}
        & Appeal to the Moon & Using the argument, “If we can put a man on the moon, we could...” as evidence for the argument.  \\
\cline{2-3}
        & Quantum Physics Fallacy & Using quantum physics in an attempt to support your claim, when in no way is your claim related to quantum physics.  \\
\cline{2-3}
        & fact to fiction fallacy & Attempting to support a narrative or argument with facts that don't support the narrative or argument.  \\
\cline{2-3}
        & Non Sequitur & Evidence or reason is irrelevant or adds very little support to the conclusion.  \\
\cline{2-3}
        & Inflation of Conflict & Reasoning that because authorities cannot agree precisely on an issue, no conclusions can be reached at all, and minimizing the credibility of the authorities, as a result.  \\
\cline{2-3}
        & Argument by Fast Talking & When fast talking is seen as intelligence and/or confidence in the truth of one’s argument; therefore, seen as evidence of the truth of the argument itself.  \\
\cline{2-3}
        & Appeal to Intuition & Evaluating an argument based on "intuition" or "gut feeling" that is unable to be articulated, rather than evaluating the argument using reason.  \\
\cline{2-3}
        & Appeal to Closure & Accepting evidence on the basis of wanting closure, or to be done with the issue.  \\
\cline{2-3}
        & Appeal to Definition & Using a dictionary’s limited definition of a term as evidence that term cannot have another meaning, expanded meaning, or even conflicting meaning.  \\
\cline{2-3}
        & Spiritual Fallacy & Insisting that something meant to be literal is actually “spiritual” as an explanation or justification for something that otherwise would not fit in an explanation.  \\
\cline{2-3}
        & gish gallop & Overwhelming an interlocutor with as many arguments as possible, without regard for accuracy or strength of the arguments.  \\
\cline{2-3}
        & Denying the Correlative & Introducing alternatives when, in fact, there are none.  \\
\cline{2-3}
        & Red Herring & Attempting to redirect the argument to another issue to which the person doing the redirecting can better respond.  \\
\cline{2-3}
        & Strawman Fallacy & Substituting a person’s actual position or argument with a distorted, exaggerated, or misrepresented version of the position of the argument.  \\
\cline{2-3}
        & Avoiding the Issue & When an arguer responds to an argument by not addressing the points of the argument.  \\
\cline{2-3}
        & Logic Chopping & Using the technical tools of logic in an unhelpful and pedantic manner by focusing on trivial details instead of directly addressing the main issue in dispute.  \\
\cline{2-3}
        & Meaningless Question & Asking a question that cannot be answered with any sort of rational meaning.  \\
\cline{2-3}
        & Failure to Elucidate & When the definition is made more difficult to understand than the word or concept being defined.  \\
\cline{2-3}
        & Argument by Gibberish & When incomprehensible jargon or plain incoherent gibberish is used to give the appearance of a strong argument, in place of evidence or valid reasons to accept the argument.  \\
\cline{2-3}
        & Hypnotic Bait and Switch & Stating several uncontroversially true statements in succession, followed by a claim that the arguer wants the audience to accept as true.  \\
\cline{2-3}
        & Traitorous Critic Fallacy & Responding to criticism by attacking a person's perceived favorability to an out-group or dislike to the in-group as the underlying reason for the criticism rather than addressing the criticism itself, and suggesting that they stay away from the issue and/or leave the in-group.  \\
\cline{2-3}
        & Having Your Cake & Making an argument, or responding to one, in such a way that it does not make it at all clear what your position is.  \\
\cline{2-3}
        & Appeal to Common Belief & When the claim that most or many people in general or of a particular group accept a belief as true is presented as evidence for the claim.  \\
\cline{2-3}
        & Appeal to Popularity & Using the popularity of a premise or proposition as evidence for its truthfulness.  \\
\cline{2-3}
        & Appeal to Common Sense & Asserting that your conclusion or facts are just “common sense” when, in fact, they are not.  \\
\cline{2-3}
        & Appeal to Common Folk & In place of evidence, attempting to establish a connection to the audience based on being a “regular person” just like each of them.  \\
\cline{2-3}
        & Appeal to Trust & The belief that if a source is considered trustworthy or untrustworthy, then any information from that source must be true or false, respectively.  \\
\cline{2-3}
        & Argument from Age & The misconception that previous generations had superior wisdom to modern man, thus conclusions that rely on this wisdom are seen accepted as true or more true than they actually are.  \\
\cline{2-3}
        & Appeal to Heaven & Asserting the conclusion must be accepted because it is the “will of God” or “the will of the gods”.  \\
\cline{2-3}
        & Appeal to Tradition & Using historical preferences of the people (tradition), either in general or as specific as the historical preferences of a single individual, as evidence that the historical preference is correct.  \\
\cline{2-3}
        & Etymological Fallacy & The assumption that the present-day meaning of a word should be/is similar to the historical meaning.  \\
\cline{2-3}
        & Genetic Fallacy & Basing the truth claim of an argument on the origin of its claims or premises.  \\
\cline{2-3}
        & Appeal to Celebrity & Accepting a claim of a celebrity based on his or her celebrity status, not on the strength of the argument.  \\
\cline{2-3}
        & Appeal to Authority & Insisting that a claim is true simply because a valid authority or expert on the issue said it was true, without any other supporting evidence offered.  \\
\cline{2-3}
        & Appeal to False Authority & Using an alleged authority as evidence in your argument when the authority is not really an authority on the facts relevant to the argument.   \\
\cline{2-3}
        & Argument from False Authority & When a person making a claim is presented as an expert who should be trusted when his or her expertise is not in the area being discussed.  \\
\cline{2-3}
        & Blind Authority Fallacy & Asserting that a proposition is true solely on the authority making the claim.  \\
\cline{2-3}
        & Argument by Personal Charm & When an argument is made stronger by the personal characteristics of the person making the argument, often referred to as “charm”.  \\
\cline{2-3}
        & Argument to the Purse & Concluding that the truth value of the argument is true or false based on the financial status of the author of the argument or the money value associated with the truth.  \\
\cline{2-3}
        & Ad Hominem Circumstantial & Suggesting that the person who is making the argument is biased or predisposed to take a particular stance, and therefore, the argument is necessarily invalid.  \\
\cline{2-3}
        & Gadarene Swine Fallacy & The assumption that because an individual is not in formation with the group, that the individual must be the one off course.  \\
\cline{2-3}
        & Ad Hominem Tu quoque & Claiming the argument is flawed by pointing out that the one making the argument is not acting consistently with the claims of the argument.  \\
\cline{2-3}
        & Bulverism & It is the assumption and assertion that an argument is flawed or false because of the arguer's characteristic associated with the arguer's identity.  \\
\cline{2-3}
        & Righteousness Fallacy & Assuming that just because a person's intentions are good, they have the truth or facts on their side.  \\
\cline{2-3}
        & Self Righteousness Fallacy & Assuming that just because your intentions are good, you have the truth or facts on your side.  \\
\cline{2-3}
        & Reductio ad Hitlerum & The attempt to make an argument analogous with Hitler or the Nazi party.  \\
\cline{2-3}
        & Ad Hominem Guilt by Association & When the source is viewed negatively because of its association with another person or group who is already viewed negatively.  \\
\cline{2-3}
        & Identity Fallacy & When one's argument is evaluated based on their physical or social identity when the strength of the argument is independent of identity.  \\
\cline{2-3}
        & Appeal to Stupidity & Attempting to get the audience to devalue reason and intellectual discourse, or devaluing reason and intellectual discourse based on the rhetoric of an arguer.  \\
\cline{2-3}
        & Ad Hominem Abusive & Attacking the person making the argument, rather than the argument itself, when the attack on the person is completely irrelevant to the argument the person is making.  \\
\cline{2-3}
        & Ad Fidentia & Attacking the person’s self-confidence in place of the argument or the evidence.  \\
\cline{2-3}
        & appeal to loyalty & When one is either implicitly or explicitly encouraged to consider loyalty when evaluating the argument when the truth of the argument is independent of loyalty.  \\
\cline{2-3}
        & Appeal to Accomplishment & When the argument being made is sheltered from criticism based on the level of accomplishment of the one making the argument.  \\
\cline{2-3}
        & Scapegoating & Unfairly blaming an unpopular person or group of people for a problem or a person or group that is an easy target for such blame.  \\
\cline{2-3}
        & Fallacy of Opposition & Asserting that those who disagree with you must be wrong and not thinking straight, primarily based on the fact that they are the opposition.  \\
\cline{2-3}
        & Proof by Intimidation & Making an argument purposely difficult to understand in an attempt to intimidate your audience into accepting it, or accepting an argument without evidence or being intimidated to question the authority or a priori assumptions of the one making the argument.  \\
\cline{2-3}
        & Poisoning the Well & To prime the audience with adverse information about the opponent from the start, in an attempt to make your claim more acceptable or discount the credibility of your opponent’s claim.  \\
\cline{2-3}
        & Wishful Thinking & When the desire for something to be true is used in place of/or as evidence for the truthfulness of the claim.  \\
\cline{2-3}
        & Appeal to Faith & It is the assertion that one must have (the right kind of) faith in order to understand the argument.  \\
\cline{2-3}
        & Notable Effort & Accepting good effort as a valid reason to accept the truth of the conclusion, even though the effort is unrelated to the truth.  \\
\cline{2-3}
        & Prejudicial Language & Loaded or emotive terms used to attach value or moral goodness to believing the proposition.  \\
\cline{2-3}
        & Special Pleading & Applying standards, principles, and/or rules to other people or circumstances, while making oneself or certain circumstances exempt from the same critical criteria, without providing adequate justification.  \\
\cline{2-3}
        & If By Whiskey & A response to a question that is contingent on the questioner’s opinions and makes use of words with strong connotations.  \\
\cline{2-3}
        & Overextended Outrage & One or more statistically rare cases are implied to be the norm or the trend (without evidence) for the purpose of expressing or inciting outrage toward an entire group.  \\
\cline{2-3}
        & Appeal to Ridicule & Presenting the argument in such a way that makes the argument look ridiculous, usually by misrepresenting the argument or the use of exaggeration.  \\
\cline{2-3}
        & Argument by Emotive Language & Substituting facts and evidence with words that stir up emotion, with the attempt to manipulate others into accepting the truth of the argument.  \\
\cline{2-3}
        & Style Over Substance & When the arguer embellishes the argument with compelling language or rhetoric, and/or visual aesthetics.  \\
\cline{2-3}
        & Appeal to Anger & When the emotions of anger, hatred, or rage are substituted for evidence in an argument.  \\
\cline{2-3}
        & Appeal to Pity & The attempt to distract from the truth of the conclusion by the use of pity.  \\
\cline{2-3}
        & Appeal to Emotion & This is the general category of many fallacies that use emotion in place of reason in order to attempt to win the argument.  \\
\cline{2-3}
        & Appeal to Flattery & When an attempt is made to win support for an argument, not by the strength of the argument, but by using flattery on those whom you want to accept your argument.  \\
\cline{2-3}
        & Appeal to Spite & Substituting spite (petty ill will or hatred with the disposition to irritate, annoy, or thwart) for evidence in an argument, or as a reason to support or reject a claim.  \\
\cline{2-3}
        & pragmatic fallacy & Claiming that something is true because the person making the claim has experienced, or is referring to someone who has experienced, some practical benefit from believing the thing to be true.  \\
\cline{2-3}
        & Appeal to Force & When force, coercion, or even a threat of force is used in place of a reason in an attempt to justify a conclusion.  \\
\cline{2-3}
        & Appeal to Fear & When fear, not based on evidence or reason, is being used as the primary motivator to get others to accept an idea, proposition, or conclusion.  \\
\hline
\multirow[t]{58}{2cm}{Informal-Insufficiency} & Fallacy of Composition & Inferring that something is true of the whole from the fact that it is true of some part of the whole.  \\
\cline{2-3}
        & Fallacy of Division & Inferring that something is true of one or more of the parts from the fact that it is true of the whole.   \\
\cline{2-3}
        & Stereotyping the fallacy & The general beliefs that we use to categorize people, objects, and events while assuming those beliefs are accurate generalizations of the whole group.  \\
\cline{2-3}
        & Ecological Fallacy & The interpretation of statistical data where inferences about the nature of individuals are deduced from inference for the group to which those individuals belong.  \\
\cline{2-3}
        & Oversimplified Cause Fallacy & When a contributing factor is assumed to be the cause, or when a complex array of causal factors is reduced to a single cause.  \\
\cline{2-3}
        & Accident Fallacy & When an attempt is made to apply a general rule to all situations when clearly there are exceptions to the rule.   \\
\cline{2-3}
        & mcnamara fallacy & When a decision is based solely on quantitative observations (i.e., metrics, hard data, statistics) and all qualitative factors are ignored.  \\
\cline{2-3}
        & Overwhelming Exception & A generalization that is technically accurate, but has one or more qualifications which eliminate so many cases that the resulting argument is significantly weaker than the arguer implies.  \\
\cline{2-3}
        & Reductio ad Absurdum & A mode of argumentation or a form of argument in which a proposition is disproven by following its implications logically to an absurd conclusion.  \\
\cline{2-3}
        & Nirvana Fallacy & Comparing a realistic solution with an idealized one, and discounting or even dismissing the realistic solution.  \\
\cline{2-3}
        & Relative Privation & Trying to make a scenario appear better or worse by comparing it to the best or worst case scenario.  \\
\cline{2-3}
        & imposter fallacy & When one suggests or claims, with insufficient evidence, that the group outliers who are viewed as damaging to the group are primarily made up of infiltrators of another group with the purpose of making the infiltrated group look bad.  \\
\cline{2-3}
        & Misleading Vividness & A small number of dramatic and vivid events are taken to outweigh a significant amount of statistical evidence.  \\
\cline{2-3}
        & Appeal to Possibility & When a conclusion is assumed not because it is probably true, but because it is possible that it is true, no matter how improbable.  \\
\cline{2-3}
        & Rights To Ought Fallacy & When one conflates a reason for one's rights (constitutional or other) with what one should do.  \\
\cline{2-3}
        & Psychogenetic Fallacy & Inferring some psychological reason why an argument is made then assuming it is invalid, as a result.  \\
\cline{2-3}
        & Weak Analogy & When an analogy is used to prove or disprove an argument, but the analogy is too dissimilar to be effective, that is, it is unlike the argument more than it is like the argument.  \\
\cline{2-3}
        & Extended Analogy & Suggesting that because two things are alike in some way and one of those things is like something else, then both things must be like that "something else".  \\
\cline{2-3}
        & Appeal to Equality & An assertion is deemed true or false based on an assumed pretense of equality, where what exactly is "equal" is not made clear, and not supported by the argument.  \\
\cline{2-3}
        & False Equivalence & An argument or claim in which two completely opposing arguments appear to be logically equivalent when in fact they are not.  \\
\cline{2-3}
        & Galileo Fallacy & The claim that because an idea is forbidden, prosecuted, detested, or otherwise mocked, it must be true, or should be given more credibility.  \\
\cline{2-3}
        & Post Designation & Drawing a conclusion from correlations observed in a given sample, but only after the sample has already been drawn, and without declaring in advance what correlations the experimenter was expecting to find.  \\
\cline{2-3}
        & Just In Case Fallacy & Making an argument based on the worst-case scenario rather than the most probable scenario, allowing fear to prevail over reason.  \\
\cline{2-3}
        & Selective Attention & Focusing your attention on certain aspects of the argument while completely ignoring or missing other parts.  \\
\cline{2-3}
        & nutpicking fallacy & When someone presents an atypical or weak member of a group as if they are a typical or strong representative.  \\
\cline{2-3}
        & Biased Sample Fallacy & Drawing a conclusion about a population based on a sample that is biased, or chosen in order to make it appear the population on average is different than it actually is.  \\
\cline{2-3}
        & Survivorship Fallacy & In its general form, the survivorship fallacy is basing a conclusion on a limited number of "winner" testimonies due to the fact we cannot or do not hear the testimonies of the losers.  \\
\cline{2-3}
        & Spotlight Fallacy & Assuming that the media’s coverage of a certain class or category is representative of the class or category in whole.  \\
\cline{2-3}
        & Hasty Generalization & Drawing a conclusion based on a small sample size, rather than looking at statistics that are much more in line with the typical or average situation.  \\
\cline{2-3}
        & Incomplete Comparison & An incomplete assertion that cannot possibly be refuted. This is popular in advertising.  \\
\cline{2-3}
        & Texas Sharpshooter Fallacy & Ignoring the difference while focusing on the similarities, thus coming to an inaccurate conclusion.  \\
\cline{2-3}
        & Faulty Comparison & Comparing one thing to another that is really not related, in order to make one thing look more or less desirable than it really is.  \\
\cline{2-3}
        & Base Rate Fallacy & Ignoring statistical information in favor of using irrelevant information, that one incorrectly believes to be relevant, to make a judgment.  \\
\cline{2-3}
        & Least Plausible Hypothesis & Choosing more unreasonable explanations for phenomena over more defensible ones.  \\
\cline{2-3}
        & Far Fetched Hypothesis & Offering a bizarre (far-fetched) hypothesis as the correct explanation without first ruling out more mundane explanations.  \\
\cline{2-3}
        & Cherry Picking & When only select evidence is presented in order to persuade the audience to accept a position, and evidence that would go against the position is withheld.  \\
\cline{2-3}
        & Argument by Selective Reading & When a series of arguments or claims is made and the opponent acts as if the weakest argument was the best one made.  \\
\cline{2-3}
        & deceptive sharing & Sharing an article, post, or meme on social media with the intent to influence public perception to perceive a statistically rare event as a common event.  \\
\cline{2-3}
        & Multiple Comparisons Fallacy & Claiming that unexpected trends that occur through random chance alone in a data set with a large number of variables are meaningful.  \\
\cline{2-3}
        & Magical Thinking & Making causal connections or correlations between two events not based on logic or evidence, but primarily based on superstition.  \\
\cline{2-3}
        & Slippery Slope & When a relatively insignificant first event is suggested to lead to a more significant event, which in turn leads to a more significant event, and so on, until some ultimate, significant event is reached, where the connection of each event is not only unwarranted but with each step it becomes more and more improbable.  \\
\cline{2-3}
        & Sunk Cost Fallacy & Reasoning that further investment is warranted on the fact that the resources already invested will be lost otherwise, not taking into consideration the overall losses involved in the further investment.  \\
\cline{2-3}
        & Jumping to Conclusions & Drawing a conclusion without taking the needed time to evaluate the evidence or reason through the argument.  \\
\cline{2-3}
        & Argument from Silence & Drawing a conclusion based on the silence of the opponent, when the opponent is refusing to give evidence for any reason.  \\
\cline{2-3}
        & Argument from Hearsay & Presenting the testimony of a source that is not an eyewitness to the event in question.    \\
\cline{2-3}
        & Anonymous Authority & When an unspecified source is used as evidence for the claim.  \\
\cline{2-3}
        & Insignificant Cause & An explanation that posits one minor factor, out of several that contributed, as its sole cause.  \\
\cline{2-3}
        & Just Because Fallacy & Refusing to respond to give reasons or evidence for a claim by stating yourself as the ultimate authority on the matter.  \\
\cline{2-3}
        & Appeal to the Law & When following the law is assumed to be the morally correct thing to do, without justification, or when breaking the law is assumed to be the morally wrong thing to do, without justification.  \\
\cline{2-3}
        & Appeal to Normality & Using social norms to determine what is good or bad.  \\
\cline{2-3}
        & False Effect & Claiming that the cause is true or false based on what we know about the effect in a claim of causality that has not been properly established.  \\
\cline{2-3}
        & Appeal to Consequences & Concluding that an idea or proposition is true or false because the consequences of it being true or false are desirable or undesirable.  \\
\cline{2-3}
        & Retrogressive Causation & Invoking the cause to eliminate the effect, or calling on the source to relieve the effect of the source.  \\
\cline{2-3}
        & Confusing Currently Unexplained with Unexplainable & Making the assumption that what cannot currently be explained is, therefore, unexplainable (impossible to explain).  \\
\cline{2-3}
        & Appeal to Desperation & Arguing that your conclusion, solution, or proposition is right based on the fact that something must be done, and your solution is "something."  \\
\cline{2-3}
        & Regression Fallacy & Ascribing a cause where none exists in situations where natural fluctuations exist while failing to account for these natural fluctuations.  \\
\cline{2-3}
        & Causal Reductionism & Assuming a single cause or reason when there were actually multiple causes or reasons.  \\
\cline{2-3}
        & Questionable Cause & Concluding that one thing caused another, simply because they are regularly associated.  \\
\hline
\multirow[t]{54}{2cm}{Informal-Inappropriate Presumption} & Hedging & Refining your claim simply to avoid counter evidence and then acting as if your revised claim is the same as the original.  \\
\cline{2-3}
        & Circular Definition & A circular definition is defining a term by using the term in the definition.  \\
\cline{2-3}
        & Homunculus Fallacy & An argument that accounts for a phenomenon in terms of the very phenomenon that it is supposed to explain, which results in an infinite regress.  \\
\cline{2-3}
        & Circular Reasoning & A type of reasoning in which the proposition is supported by the premises, which is supported by the proposition.  \\
\cline{2-3}
        & Tokenism & Interpreting a token gesture as an adequate substitute for the real thing.  \\
\cline{2-3}
        & Appeal to Novelty & Claiming that something that is new or modern is superior to the status quo, based exclusively on its newness.  \\
\cline{2-3}
        & Two Wrongs Make a Right & When a person attempts to justify an action against another person because the other person did take or would take the same action against him or her.  \\
\cline{2-3}
        & Appeal to Nature & When used as a fallacy, the belief or suggestion that “natural” is better than “unnatural” based on its naturalness.  \\
\cline{2-3}
        & Naturalistic Fallacy & When the conclusion expresses what ought to be, based only on what is, or what ought not to be, based on what is not.  \\
\cline{2-3}
        & Moralistic Fallacy & When the conclusion expresses what is, based only on what one believes ought to be, or what isn’t is based on what one believes ought not to be.  \\
\cline{2-3}
        & Suppressed Correlative & The attempt to redefine a correlative (one of two mutually exclusive options) so that one alternative encompasses the other, i.e. making one alternative impossible.  \\
\cline{2-3}
        & Historians Fallacy & Judging a person's decision in the light of new information not available at the time.  \\
\cline{2-3}
        & Willed Ignorance & Refusing to change one’s mind or consider conflicting information based on a desire to maintain one's existing beliefs.  \\
\cline{2-3}
        & Appeal to Coincidence & Concluding that a result is due to chance when the evidence strongly suggests otherwise.  \\
\cline{2-3}
        & Argument from Incredulity & Concluding that because you can't or refuse to believe something, it must not be true, improbable, or the argument must be flawed.  \\
\cline{2-3}
        & Argument by Pigheadedness & This is a refusal to accept a well-proven argument for one of many reasons related to stubbornness.  \\
\cline{2-3}
        & Argument by Repetition & Repeating an argument or a premise over and over again in place of better supporting evidence.  \\
\cline{2-3}
        & Definist Fallacy & Defining a term in such a way that makes one’s position much easier to defend.  \\
\cline{2-3}
        & Limited Scope & The theory doesn't explain anything other than the phenomenon it explains (that one thing), and at best, is likely to be incomplete.  \\
\cline{2-3}
        & Moving the Goalposts & Demanding from an opponent that he or she address more and more points after the initial counter-argument has been satisfied refusing to concede or accept the opponent’s argument.  \\
\cline{2-3}
        & Argument from Fallacy & Concluding that the truth value of an argument is false based on the fact that the argument contains a fallacy.  \\
\cline{2-3}
        & False Dilemma & When only two choices are presented yet more exist, or a spectrum of possible choices exists between two extremes.  \\
\cline{2-3}
        & Argument from Ignorance & The assumption of a conclusion or fact based primarily on lack of evidence to the contrary.  \\
\cline{2-3}
        & Alternative Advance & When one is presented with just two choices, both of which are essentially the same, just worded differently.  \\
\cline{2-3}
        & Shifting of the Burden of Proof & Making a claim that needs justification, then demanding that the opponent justifies the opposite of the claim.  \\
\cline{2-3}
        & Proving Non Existence & Demanding that one proves the non-existence of something in place of providing adequate evidence for the existence of that something.  \\
\cline{2-3}
        & Proof Surrogate & A claim masquerading as proof or evidence, when no such proof or evidence is actually being offered.  \\
\cline{2-3}
        & Rationalization & Offering false or inauthentic excuses for our claim because we know the real reasons are much less persuasive or more embarrassing to share, or harsher than the manufactured ones given.  \\
\cline{2-3}
        & Spin Doctoring & Presenting information in a deceptive way that results in others interpreting the information in such a way that does not reflect reality but is how you want the information to be interpreted.  \\
\cline{2-3}
        & Lying with Statistics & Presenting statistical data in a very biased way.  \\
\cline{2-3}
        & Ad Hoc Rescue & When we desperately want to be right and hold on to certain beliefs, we begin to make up excuses without no real evidence as to why our belief could be.  \\
\cline{2-3}
        & False Attribution & Appealing to an irrelevant, unqualified, unidentified, biased, or fabricated source in support of an argument (modern usage).  \\
\cline{2-3}
        & Amazing Familiarity & The argument contains information that seems impossible to have obtained—like it came from an omniscient author.  \\
\cline{2-3}
        & Ludic Fallacy & Assuming flawless statistical models apply to situations where they actually don’t.   \\
\cline{2-3}
        & Missing Data Fallacy & Refusing to admit ignorance to the hypothesis and/or the conclusion, but insisting that your ignorance has to do with missing data that validate both the hypothesis and conclusion.  \\
\cline{2-3}
        & Begging the Question & Any form of argument where the conclusion is assumed in one of the premises.  \\
\cline{2-3}
        & Complex Question Fallacy & A question that has a presupposition built in, which implies something but protects the one asking the question from accusations of false claims.  \\
\cline{2-3}
        & Package Deal Fallacy & Assuming things that are often grouped together must always be grouped together, or the assumption that the ungrouping will have significantly more severe effects than anticipated.  \\
\cline{2-3}
        & Subjectivist Fallacy & Claiming something is true for one person, but not for someone else when, in fact, it is true for everyone (objective) as demonstrated by empirical evidence.  \\
\cline{2-3}
        & Distinction Without a Difference & The assertion that a position is different from another position based on the language when, in fact, both positions are the same -- at least in practice or practical terms.  \\
\cline{2-3}
        & Hypothesis Contrary to Fact & Offering a poorly supported claim about what might have happened in the past or future, if (the hypothetical part) circumstances or conditions were different.  \\
\cline{2-3}
        & Shoehorning & The process of force-fitting some current affair into one's personal, political, or religious agenda.  \\
\cline{2-3}
        & Appeal to Self-evident Truth & Making the claim that something is "self-evident" when it is not self-evident in place of arguing a claim with reason.  \\
\cline{2-3}
        & Subverted Support & The attempt to explain some phenomenon that does not actually occur or there is no evidence that it does.  \\
\cline{2-3}
        & Double Standard & Judging two situations by different standards when, in fact, you should be using the same standard.  \\
\cline{2-3}
        & Fantasy Projection & Confusing subjective experiences, usually very emotionally charged, with objective reality, then suggesting or demanding that others accept the subjective experience as objective reality.  \\
\cline{2-3}
        & Argument to Moderation & Asserting that given any two positions, there exists a compromise between them that must be correct.  \\
\cline{2-3}
        & Broken Window Fallacy & The illusion that destruction and money spent in recovery from destruction, is a net-benefit to society.  \\
\cline{2-3}
        & Self Sealing Argument & An argument or position is self-sealing if and only if no evidence can be brought against it no matter what.  \\
\cline{2-3}
        & Unfalsifiability & Confidently asserting that a theory or hypothesis is true or false even though the theory or hypothesis cannot possibly be contradicted by an observation or the outcome of any physical experiment, usually without strong evidence or good reasons.  \\
\cline{2-3}
        & Conspiracy Theory & Explaining that your claim cannot be proven or verified because the truth is being hidden and/or evidence destroyed by a group of two or more people.  \\
\cline{2-3}
        & Confusing an Explanation with an Excuse & Treating an explanation of a fact as if it were a justification of the fact, a valid reason for the fact, or evidence for the fact.  \\
\cline{2-3}
        & Limited Depth & Failing to appeal to an underlying cause, and instead simply appealing to membership in a category.  \\
\cline{2-3}
        & Alleged Certainty & Asserting a conclusion without evidence or premises, through a statement that makes the conclusion appear certain when, in fact, it is not.  \\
\hline
\caption{Categorizations and descriptions of the fallacies in our dataset.}
\label{tab:fallacy_232}
\end{longtable}
\end{center}

}

\clearpage
\twocolumn

\end{document}